\documentclass[sn-mathphys-num]{sn-jnl}%
\usepackage{graphicx}%
\usepackage{mathrsfs} 
\usepackage{multirow}%
\usepackage{amsmath,amssymb,amsfonts}%
\usepackage{amsthm}%
\usepackage{mathrsfs}%
\usepackage{float}
\usepackage[title]{appendix}%
\usepackage{xcolor}%
\usepackage{textcomp}%
\usepackage{manyfoot}%
\usepackage{booktabs}%
\usepackage{algorithm}%
\usepackage{algorithmicx}%
\usepackage{algpseudocode}%
\usepackage{listings}%
\usepackage{tikz}
\usepackage{graphicx}
\usetikzlibrary{shapes, arrows, positioning}
\usepackage{svg}
\usepackage{algorithm}
\usepackage{algpseudocode}
\usepackage{amsmath}
\usepackage{cleveref}

\theoremstyle{thmstyleone}%

\theoremstyle{thmstyletwo}%

\theoremstyle{thmstylethree}%

\begin{document}
\title[Article Title]{Hybrid Adaptive Modeling in Process Monitoring: Leveraging Sequence Encoders and Physics-Informed Neural Networks}

\author*[1,2]{\fnm{Mouad} \sur{ELAARABI}}

\author[1]{\fnm{Domenico} \sur{BORZACCHIELLO}}

\author[2]{\fnm{Philippe} \sur{LE BOT}}

\author[2]{\fnm{Nathan} \sur{LAUZERAL}}

\author[1]{\fnm{Sebastien} \sur{COMAS-CARDONA}}

\affil[1]{\orgdiv{Nantes Université}, \orgname{Ecole Centrale Nantes, CNRS, GeM, UMR 6183}, \orgaddress{\city{Nantes}, \postcode{44321}, \country{France}}}

\affil[2]{\orgdiv{Nantes Université}, \orgname{IRT Jules Verne}, \orgaddress{\city{Bouguenais}, \postcode{44340}, \country{France}}}

\abstract{In this work, we explore the integration of Sequence-Encoding for Online Parameter Identification with Physics-Informed Neural Networks (PINNs) to create a model that, once trained, can be utilized for real-time applications with variable parameters, boundary conditions (BCs), and initial conditions (ICs). Recently, the combination of PINNs with Sparse Regression (PINN-SR) has emerged as a method for performing dynamical system identification through supervised learning and sparse regression optimization, while also solving the dynamics using PINNs. However, this approach can be limited by variations in parameters or boundary/initial conditions, requiring retraining of the model whenever changes occur. 
In this work, we introduce an architecture that employs Deep Sets (Sequence Encoders) to encode dynamic parameters, boundary conditions, and initial conditions, using these encoded features as inputs for the PINN, enabling the model to adapt to changes in parameters, BCs, and ICs. \\
We apply this approach to three different problems: first, we analyze the Rössler ODE system, demonstrating the robustness of the model with respect to noise and its ability to generalize. Next, we explore the model's capability in a 2D Navier-Stokes PDE problem involving flow past a cylinder with a parametric sinusoidal inlet velocity function, showing that the model can encode pressure data from a few points to identify the inlet velocity profile and utilize physics to compute velocity and pressure throughout the domain. Finally, we address a 1D heat monitoring problem using real data from the heating of glass fiber and thermoplastic composite plates.
 }


\maketitle

\section{Introduction}

Machine learning methods, especially deep learning techniques, are receiving significant attention across various research fields due to their ability to learn complex relationships between input and output features. However, these methods often suffer from a lack of interpretability, as they function as black-box models, making it difficult to extract meaningful insights from their predictions. Additionally, they generally require extensive training data to achieve high accuracy, which may not always be readily available. This limitation affects their ability to generalize well to unseen data~\cite{rudin2019stop}. While techniques such as Neural Tangent Kernels~\cite{jacot2018neural} (NTK) can be employed to analyze the behavior of neural networks, they do not fully resolve these challenges.

The rapid growth of research in physics-informed machine learning~\cite{PIML} (PIML) has led to innovative methods that integrate physics-based knowledge into data-driven models~\cite{chao2022fusing, willard2020integrating, xu2023physics}. These approaches benefit from advancements in both machine learning algorithms and hardware acceleration, making them increasingly accessible. Unlike conventional neural networks, physics-informed models incorporate governing physical laws into the learning process, improving their ability to generalize beyond available supervised data. A notable example is Physics-Informed Neural Networks (PINNs), which leverage both physics-based constraints and observed data to solve Partial and Ordinary Differential Equations (PDEs/ODEs). PINNs have gained considerable interest due to their successful applications in direct problems, in various fields such fluid dynamics~\cite{yang2022learning}, multiphysics coupled problems~\cite{ma2022preliminary, coulaud2023physics}, as well as their ability to handle multi-fidelity data sources~\cite{PINNMF}. PINNs could also be used for inverse problems~\cite{raissi2019physics,hanna2024self, boulenc2025spatially}. They have also been benchmarked against Finite Element Methods (FEM)~\cite{DEEPXDE}, demonstrating very good performance. Furthermore, PINNs have shown robustness in solving inverse problems, effectively handling noisy data to infer accurate underlying dynamics. This ability to explore physics through data-driven methods has become a highly active research area.

The direct derivation of governing equations from first principles can be highly complex, as it is sensitive to parameter variations and modeling assumptions. In these cases, researchers can rely on data-driven approaches to extract system dynamics through empirical data analysis, simulations, or experiments. One particularly promising direction in this field is the identification of dynamic systems from empirical data, where the goal is to extract governing equations (as differential equations PDEs/ODEs) that describe interactions between state variables using concise mathematical expressions. Deriving these equations solely from theoretical principles is often challenging due to the complexity of real world systems. However, once identified, these equations can be applied in PINN-based simulations, FEM analyses, and industrial process monitoring and control~\cite{SINDYMPC}. A widely recognized approach in this context is Sparse Identification of Nonlinear Dynamics (SINDy~\cite{SINDy}), which employs sparse optimization techniques to identify the most relevant terms from a predefined search library that accurately describe system dynamics.

\subsection{Background and motivations}
Recently, a novel approach called PINN-SR~\cite{PINNSR} proposed combining PINNs and SINDy to simultaneously train a model for both identification and solving PDEs using scarce and noisy data. This method was tested on well-known PDEs, including the Navier-Stokes equations, Burgers' equation, and other dynamics, demonstrating good robustness to noise and high accuracy despite scarce and noisy data. This approach reduces training time and leverages automatic differentiation to generate the search library that combines partial derivatives with respect to the input data, making it a promising solution for real-world applications where generating large, high-quality datasets is challenging.\\

However, PINN-SR remains constrained by its need for retraining when system parameters or boundary conditions change, or in cases where parameters have time/space dependencies. To address this limitation, we recently proposed in~\cite{SEQENCODER} to generalize SINDy using Sequence Encoders. Two types of architectures, Deep sets~\cite{DEEPSETS} and Set Transformer~\cite{SETTRANS, ATTTRF}, have been challenged for online identification, enabling real time adaptation to dynamic systems. By incorporating Deep sets and Set Transformers, with PINN advantages, we extended PINN-SR to handle variable parameters, with time/space dependencies, initial conditions, and boundary conditions without requiring full model retraining. This advancement described in~\cite{SEQENCODER} paves the way for more efficient, real-time physics-informed learning frameworks, applicable to a wide range of dynamics problems.\\

Recently, Deep Operator Sets~\cite{DeepOSets} (DeepOSets) proposed combining Deep Operator Network~\cite{DeepONets} (DeepONet), a data-driven model trained to learn mappings between function spaces, with the Deep sets architecture. DeepONet consists of a branch network that encodes sensor data and a trunk network that encodes the target domain (query points). The final output is obtained through the dot product of the outputs of the two networks. Once trained, this model is used to encode sensor data and solve parametric partial differential equations using feed-forward prediction. DeepOSets enhances this model by integrating Deep sets, enabling DeepONet to generalize for variable-length snapshots and their permutations, and to be more robust to noise.

\subsection{Paper contributions}
In this paper, we introduce an approach to generalize PINN-SR, to have a unique model that does not require retraining each time we have changes in parameters and BCs/ICs. Hence, this approach is different from existing methods such as Physics-Informed DeepONets, as we use multiple blocks for parameter identification, BCs/ICs approximation, and then these blocks are trained together with PINN. This approach will allow us to perform cross-validation of block predictions to provide some level of confidence in their predictions. Additionally, the training process is easier to manipulate, as each block can be pre-trained separately. Finally, we do not aim to train a universal model that takes into consideration all BCs/ICs or parameter changes, but only changes within a predefined domain. Moreover, the proposed model can encode variable sequence lengths (time series from sensors), as demonstrated in~\cite{SEQENCODER}. The Sequence Encoder, which is the encoder block of our model, can be fixed in terms of the number of input sensors (as in the case of this article) or dynamic, as we will explain later, since the model architecture allows for the incorporation of variable sensors. The model can also take other information as input, such as dimensions and BC/ICs. The problem we are addressing in this article can be described as follows Eqs.~\ref{intro.eq.1}:
\begin{equation}\label{intro.eq.1}
\begin{aligned}
    F\left(x_1, x_2, \dots, x_m, \mathbf{u}, \frac{\partial \mathbf{u}}{\partial x_1}, \frac{\partial \mathbf{u}}{\partial x_2}, \dots, \frac{\partial^2 \mathbf{u}}{\partial x_i \partial x_j}, \dots \right) &= \mathbf{0}, \\
    \mathbf{u} &= S(\mathbf{f}, \mathbf{g})
\end{aligned}
\end{equation}
where the first equation represents a general form of a PDE/ODE, where $x_i$ are the independent variables, $\textbf{u}$ are the state variables, and $F$ represents the system of PDEs/ODEs, and the second is the expression of the solution operator $S$ in a predefined domain of variation of vectors $\mathbf{f}$, which represents the force information, and $\mathbf{g}$, which represents BC/ICs, and the dynamics parameters. In this work, we used sensor data and BC/ICs as input to the Sequence Encoder, which simplifies to Eqs.~\ref{intro.eq.2} : 
\begin{equation}\label{intro.eq.2}
\begin{aligned}
    \mathbf{SE}(s_0, \dots s_i, \dots s_p, ~~BCs, ~~ICs) &= \mathbf{F_{vec}}, \\
    \mathbf{u} &= \mathbf{PINN}(x_0, \dots x_i, \dots x_n;~~\mathbf{F_{vec}})
\end{aligned}
\end{equation}
where $\mathbf{SE}$ is the Sequence Encoder, $\mathbf{F_{vec}}$ is the feature vector, and PINN is a trainable Multi Layer Perceptron (MLP) with nonlinear activation functions, and $s_i$ are the sensors. The contributions of this paper are as follows:
\begin{itemize}
    \item The combination of PINN and Sequence Encoder.
    \item Generalization of PINN-SR to handle variable BC/ICs and parameters, using the Sequence Encoder to encode sensor data with variable lengths.
    \item The robustness of the model to noisy and scarce data, and the cases where the frequency of acquisition (or $\Delta T$) is variable.
    \item Cross-validation of the model's predictions through multi-block predictions.
\end{itemize}

\subsection{Outline of the Paper}\label{sec_rep}
This paper is organized as follows. First, we briefly introduce the state of the art methods SINDy and PINN, and their combination PINN-SR in Sec.~\ref{SINDy-PINN}. We then provide a brief description of Sequence Encoding methods, Set Transformer and Deep-Set, in Sec.~\ref{SE}. Following that, we give an overall description of the architecture proposed in this paper in Sec.~\ref{PINN-SE}. The model is validated using three applications, as described in Sec.~\ref{Appli_into}. In the first application (Sec.~\ref{ODE-App}), we study the model's ability to solve the Rossler ODE with variable ICs/parameters near its chaotic behavior, based on scarce data (time series) with variable lengths, as well as the model's robustness to noise. In the next application (Sec.~\ref{NS-App}), we study a the problem of 2D incompressible flow around a cylinder with time-varying parametric boundary conditions. In this application, we provide only pressure data at limited points to the sequence encoder to predict the inlet function and also predict the velocity using physics in the entire domain. The last application (Sec.~\ref{HT-App}), tests the model on acquired experimental data from heating thermoplastic composite materials during the heating process. The problem is a 1D heat transfer problem with variable material thickness, thermal properties, and heat flux BCs, as well as sudden changes in BCs.

\section{Identification methods and PINN}\label{SINDy-PINN}
Symbolic regression and genetic algorithms are widely used for extracting mathematical expressions from input-output data. These methods can be extended to learn PDEs/ODEs by incorporating differentiation operators within genetic algorithms, as demonstrated by \cite{SAIND}, or by using convolution kernels to represent such operators, as shown by \cite{PDENET}.

\begin{figure}[!h]
    \centering
    
    \begin{tikzpicture}

        \tikzstyle{block} = [rectangle, rounded corners, draw=black, fill=white, text=black, minimum width=3.5cm, minimum height=1cm, align=center]
        \tikzstyle{mainblock} = [rectangle, draw=black, fill=red, text=white, minimum width=3.5cm, minimum height=1cm, align=center]
        \tikzstyle{arrow} = [thick,->,>=stealth]

        \node (input) [block] {Dynamics state\\ variables, search library};
        \node (framework) [mainblock, right=0.5cm of input] {SINDy\\ Framework\cite{PySINDy}};
        \node (output) [block, right=0.5cm of framework] {System\\ parameters};
        \node (optimization) [block, below=0.5cm of framework] {Optimization algorithm\\ (LASSO~\cite{LASSO}, SR3~\cite{SR3}, $\cdots$)\\ and the hyper-parameters\\ (L0 threshold, L1 $\cdots$)};
        
        \draw [arrow] (input) -- (framework);
        \draw [arrow] (framework) -- (output);
        \draw [arrow] (optimization) -- (framework);
    \end{tikzpicture}
    \label{SINDy_Framework}
    \caption{Overview of the Sparse Identification of Nonlinear Dynamical Systems (SINDy) framework.}
\end{figure}
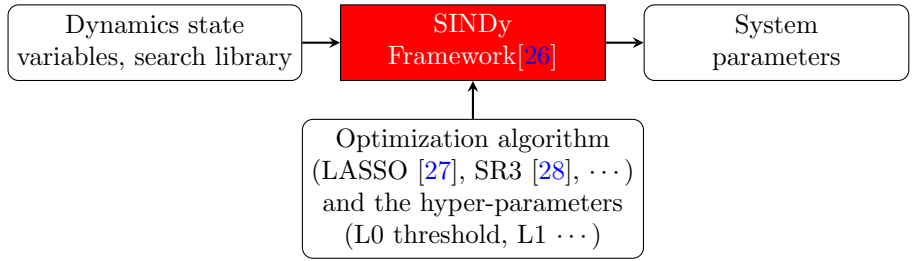

SINDy (Fig.~\ref{SINDy_Framework}) is an example of this methods. While PDEs/ODEs describe the relationships between state variables, the user defines a search library, \(\Theta(X)\) (an example of which is given in Eq.~\ref{eq.5}), consisting of candidate functions based on prior knowledge of the system's dynamics.
 
\begin{equation}\label{eq.5}
    \Theta(X) = [1~~X^T~~(X^{T})^2~~\cdots~~\sin(X^T)~~\cdots \exp(X^T)~~\cdots]
\end{equation}
Then, by employing sparse optimization techniques such as Sparse Relaxed Regularized Regression (SR3)~\cite{SR3}, or Sequential Threshold Ridge regression (STRidge)~\cite{STRIdge}, a subset of the most relevant terms is selected to accurately describe the system's dynamics. This can be formulated as Eqs.~\ref{eq.6}:\\

\begin{equation}\label{eq.6}
\begin{aligned}
    \dot{X} = \quad  \quad\Theta(X) \Xi \\
    \underset{\Xi}{\text{argmin}} \left( \sum_{i=1}^{N} \left| \dot{X}_i - \Theta(X_i) \Xi \right|^2 \right) \\
    \
\end{aligned}
\end{equation}
where \(\Xi\) represents the structure and parameters of the dynamics. In this same context, PINN could be applied for inverse problems to approximate the dynamics parameters in cases where the form of the dynamics is well known.

\begin{figure}[!th]
    \centering
    \includegraphics[width=0.7\textwidth]{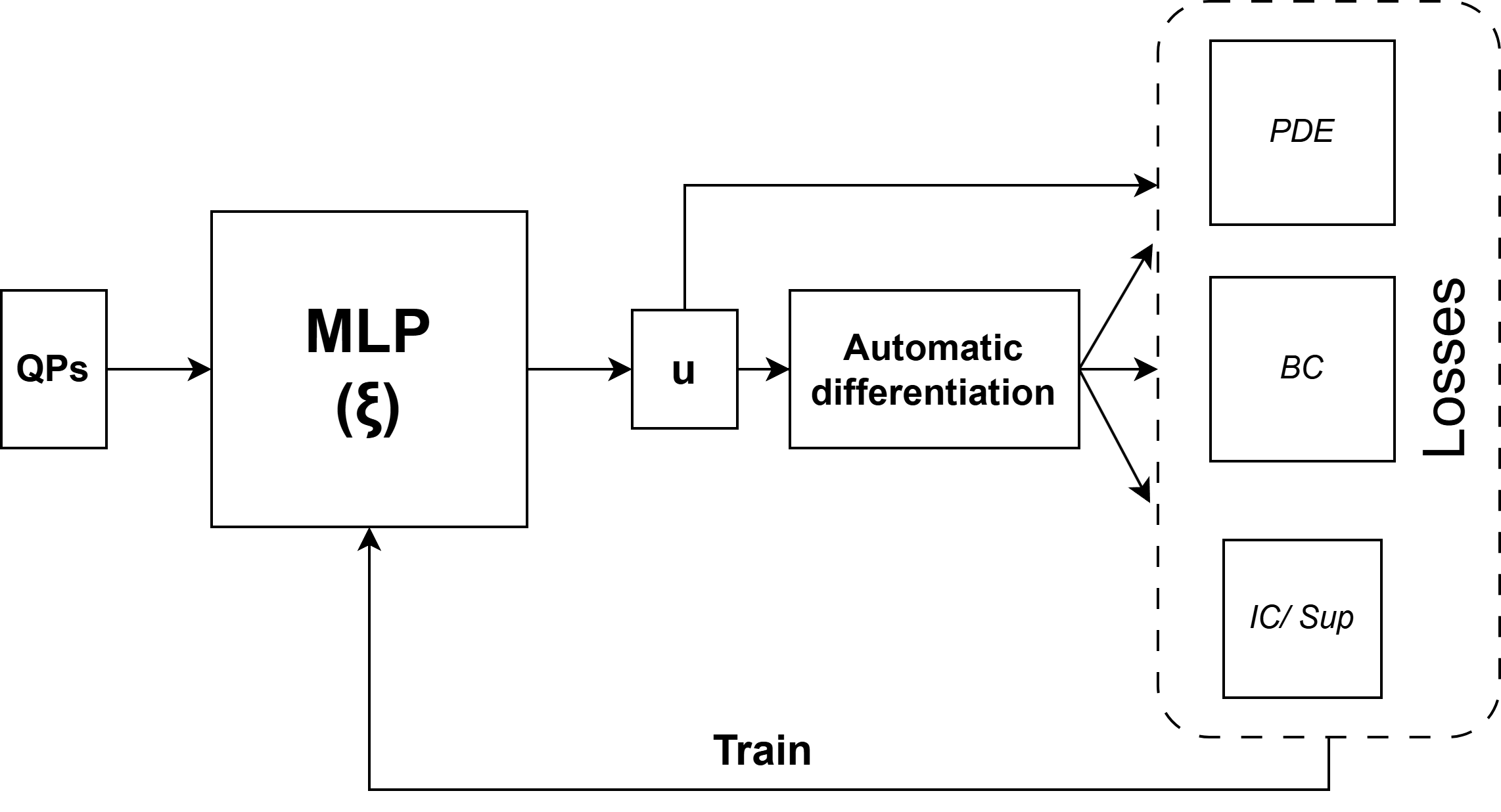}
    \caption{Example of the PINN model: the neural network learns (Multi Layer Perceptron) to solve PDEs/ODEs by minimizing a loss function that incorporates the governing equations, boundary/initial conditions (BC/IC), and supervised data. Here QPs represent the Query Points, which could correspond to time or a time and space; $\mathbf{u}$ denotes the predictions of the model; and $\xi$ are the trainable parameters of the PINN.}
    \label{PINN-fig}
\end{figure}

\begin{figure}[!h]
    \centering
    \includegraphics[width=0.7\textwidth]{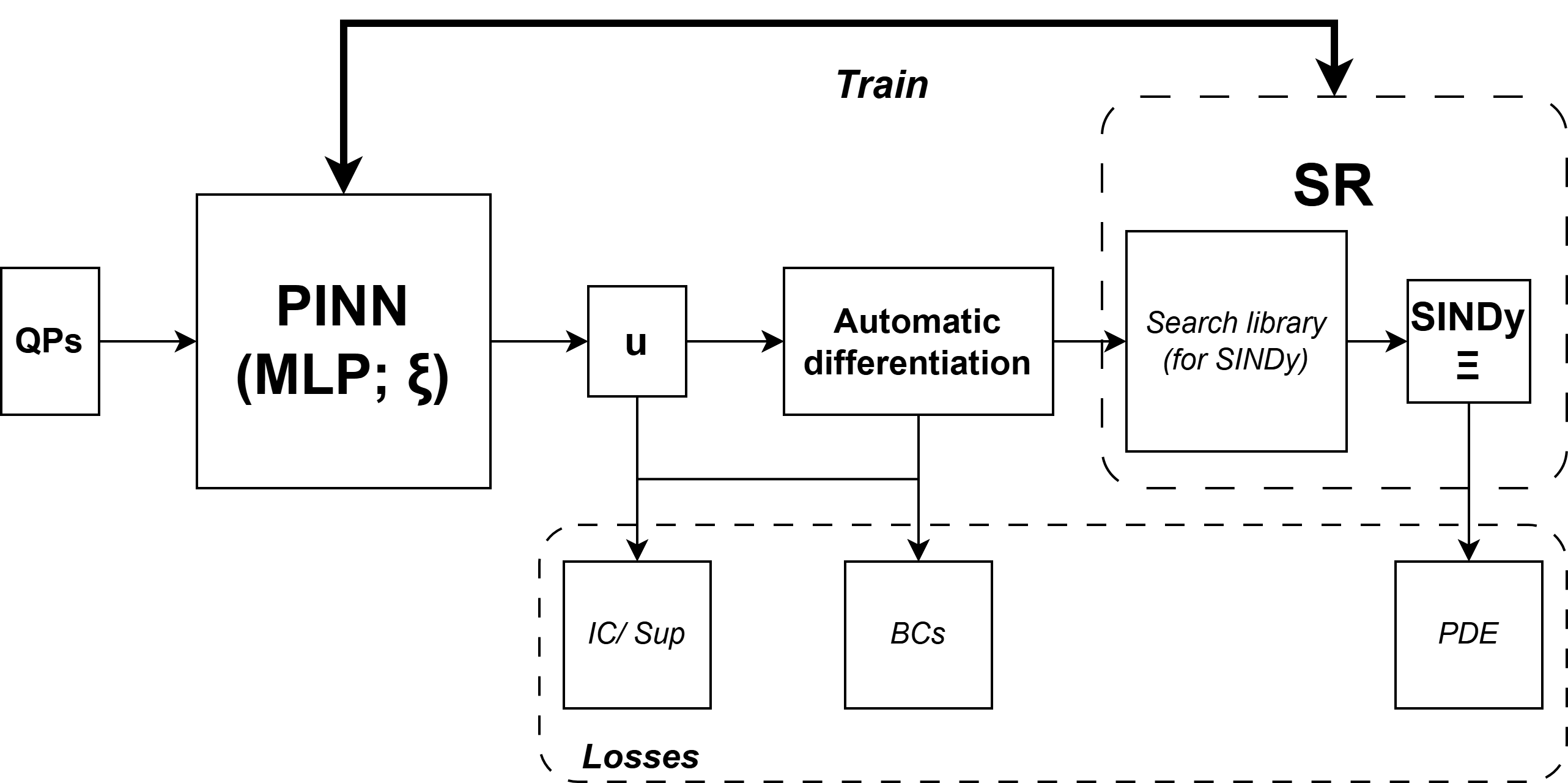}
    \caption{Overview of the PINN-SR model: a combination of PINN and SINDy to learn governing equations and solve system dynamics, where $\xi$ are the trainable variables of the PINN and $\Xi$ are the dynamic parameters to be identified. The SINDy search library is constructed using the predictions of the model $\mathbf{u}$ and its partial derivatives with respect to the query points (QPs), which could be for example, time or time and space. SINDy is then used to calculate the dynamics loss (represented as the \textit{PDE} block in the figure).}
    \label{PINN-SR-fig}
\end{figure}

PINN can also be used for forward problems to solve PDE/ODE (Fig.~\ref{PINN-fig}). The advantages of PINN come from leveraging the automatic differentiation of neural networks, which makes it easy to compute partial derivatives. This allows for the construction of the PDE/ODE and incorporating it into the loss function as training information by minimizing the following Eq.~\ref{eq.7}, where $\xi$ represents the trainable variables of PINN : 
\begin{equation}\label{eq.7}
\underset{\xi}{\text{argmin}} \quad \lambda \mathcal{L}_{\text{BC}}(\xi) + \mathcal{L}_{\text{IC/ SUP}}(\xi) + \beta \mathcal{L}_{\text{PDE}}(\xi) 
\end{equation}
Where $\mathcal{L}_{\text{BC}}$, $\mathcal{L}_{\text{IC/SUP}}$, and $\mathcal{L}_{\text{PDE}}$ represent the boundary condition loss, the initial condition and supervised loss, and the partial differential equation loss, respectively.
PINN-SR proposes to combine both SINDy and PINN for cases where limited and low-quality data are available to identify and solve PDEs from scarce, noisy, and labeled data. The architecture of PINN-SR described in Fig.~\ref{PINN-SR-fig}, is divided into two blocks, MLP that predicts the state variables and calculates the dynamics terms to construct a search library. The model is then trained to minimize the SINDy loss (physics loss), the supervised labeled data loss, and the BC/IC losses. The SINDy loss is based on Least Absolute Shrinkage and Selection Operator (LASSO)~\citep{LASSO}, which uses the residual from the first equation of Eq.~\ref{eq.6} with an \(\ell_1\)-norm error \(\left\|\Xi \right\|_1\), to minimize the number of selected terms (imposing sparsity). At the end of training, an additional step using STRidge~\cite{STRIdge} could be applied to minimize the number of terms by employing the \(\ell_0\)-norm, \(\left\|\Xi \right\|_0\). This approach is promising and is generalizable to cases where BCs/ICs, or parameters change, or when there are state, space, or time dependencies within the parameters. These variations can be handled by generalizing the SINDy framework. In the next sections~\ref{SE} and~\ref{PINN-SE}, we present an approach to generalize SINDy and combine it with PINN.

\section{Sequence-Encoding}\label{SE}
\begin{figure}[!h]
    \centering
    \includegraphics[width=0.7\textwidth]{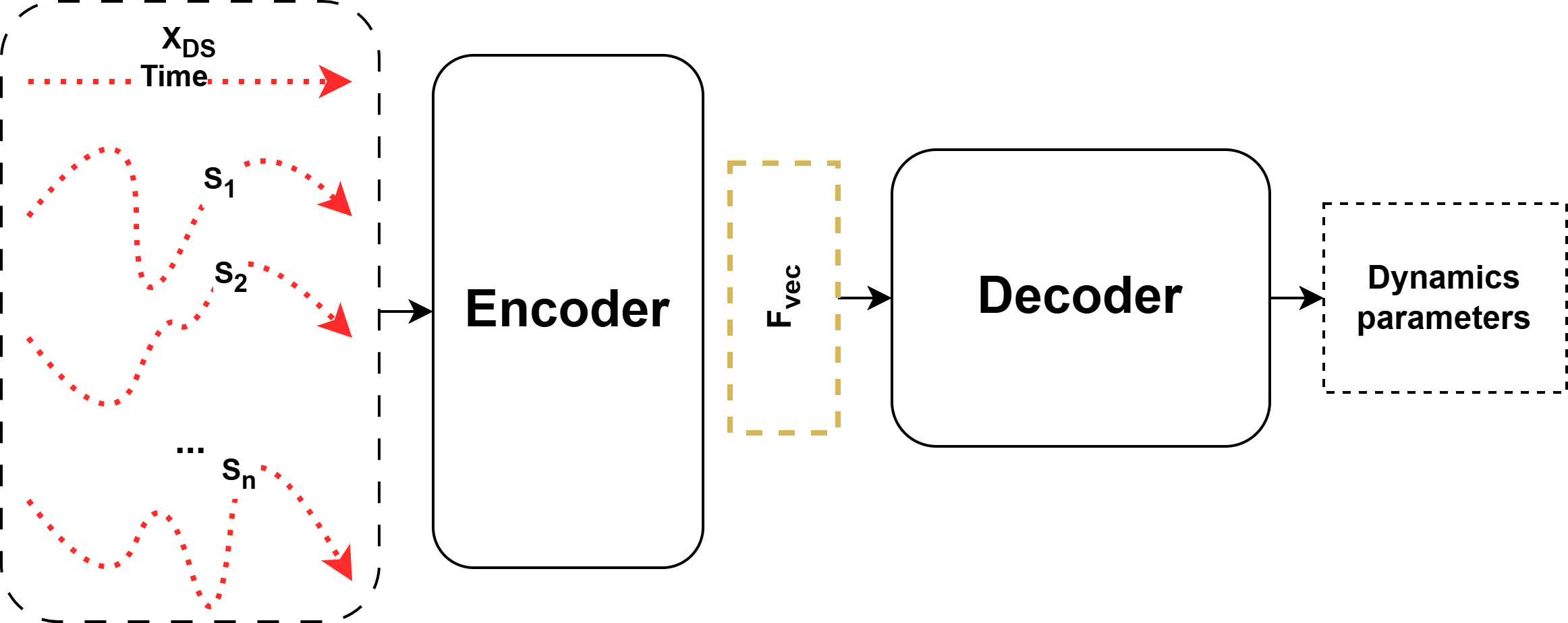}
    \caption{Overview of the Sequence Encoder (Deep Sets or Set Transformer) model blocks for capturing dynamic parameters based on sensor data. Let $s_i$ represent the sensor data. The encoder input is defined as $X_{DS} = [time,~s_1,~\cdots,~s_j,~\cdots,~s_n] \in \mathbb{R}^{\text{b} \times N \times (n+1)}$. It then generates a feature vector $\mathbf{F}_{\text{vect}} \in \mathbb{R}^{\text{b} \times n_f}$, where $n_f$ is the dimension of the feature vector. This vector is subsequently decoded to compute the dynamic parameters.}
    \label{SequenceEnc-fig}
\end{figure}

In this section, we briefly introduce sequence encoding methods. Recently, we introduced the use of these methods~\cite{SEQENCODER} for time series encoding to capture dynamic features through state variables and predict dynamic parameters. These methods (Fig.~\ref{SequenceEnc-fig}) take as input the state variables (or a set of functions/ search library $\Theta$). The encoding part differs from Deep sets, which use a simple MLP to project the input features to a higher dimension, while the Set Transformer uses self-attention blocks to encode the dynamic behavior and the interactions between the state variables, providing additional advantages. The next block is a pooling layer over the time dimension, which produces a time-independent feature vector. The pooling used by Deep sets are aggregation functions ($\text{mean},  \text{sum}, \text{max}, \dots$), while the Set Transformer uses a PMA block (Pooling Multihead Attention) with a trainable seed vector. This means that the pooling is trainable and does not assign equal importance to each time point (as in the cases of max, mean, etc.). The last block is the decoding/mapping block that converts the feature vector into the dynamic parameters. \\

\begin{figure}[!h]
    \centering
    \includegraphics[width=0.7\textwidth]{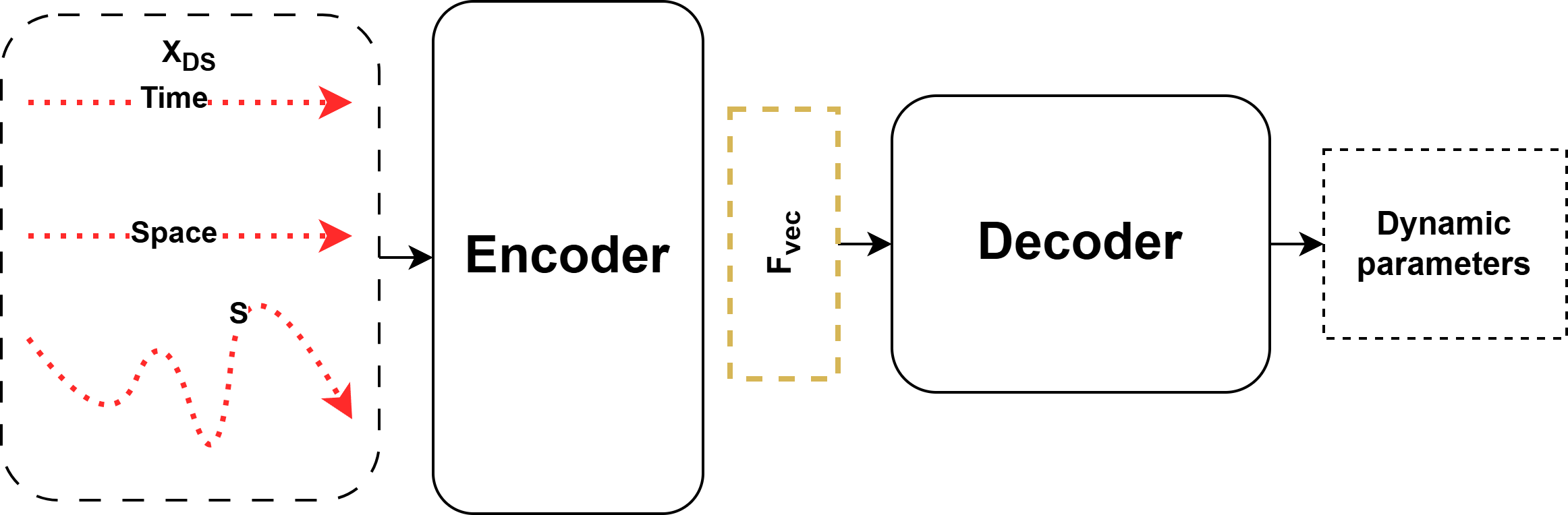}
    \caption{Overview of the Sequence Encoder (Deep sets or Set Transformer) model blocs for capturing dynamic parameters where the number and positions of the sensors are variable and dynamic. The encoder input is defined as $X_{DS} = [time,~space,~s] \in \mathbb{R}^{\text{b} \times N \times 3}$}
    \label{SequenceEnc-fig-DS}
\end{figure}

Using this encoding block to capture dynamic changes based on sensor data and BC/IC is the fundamental idea of our method. While in this application, the number of sensors is fixed (sensor locations), we believe that these methods could also be applied to cases where the number of sensors is variable and their positions are dynamic, by adding spatial information as input to the Sequence Encoder as described in Fig.~\ref{SequenceEnc-fig-DS}. In the next section~\ref{PINN-SE}, we present the combination of Deep sets as an encoder, PINN, and other identification blocks. In this paper, we focus only on Deep Sets, simply because it has fewer hyperparameters and is easier to implement; however, the Set Transformer could also be used as an encoder.

\section{Architecture of the combination of Sequence Encoder and Physics Informer Neural Networks}\label{PINN-SE}
\begin{figure}[!h]
    \centering
    \includegraphics[width=0.8\textwidth]{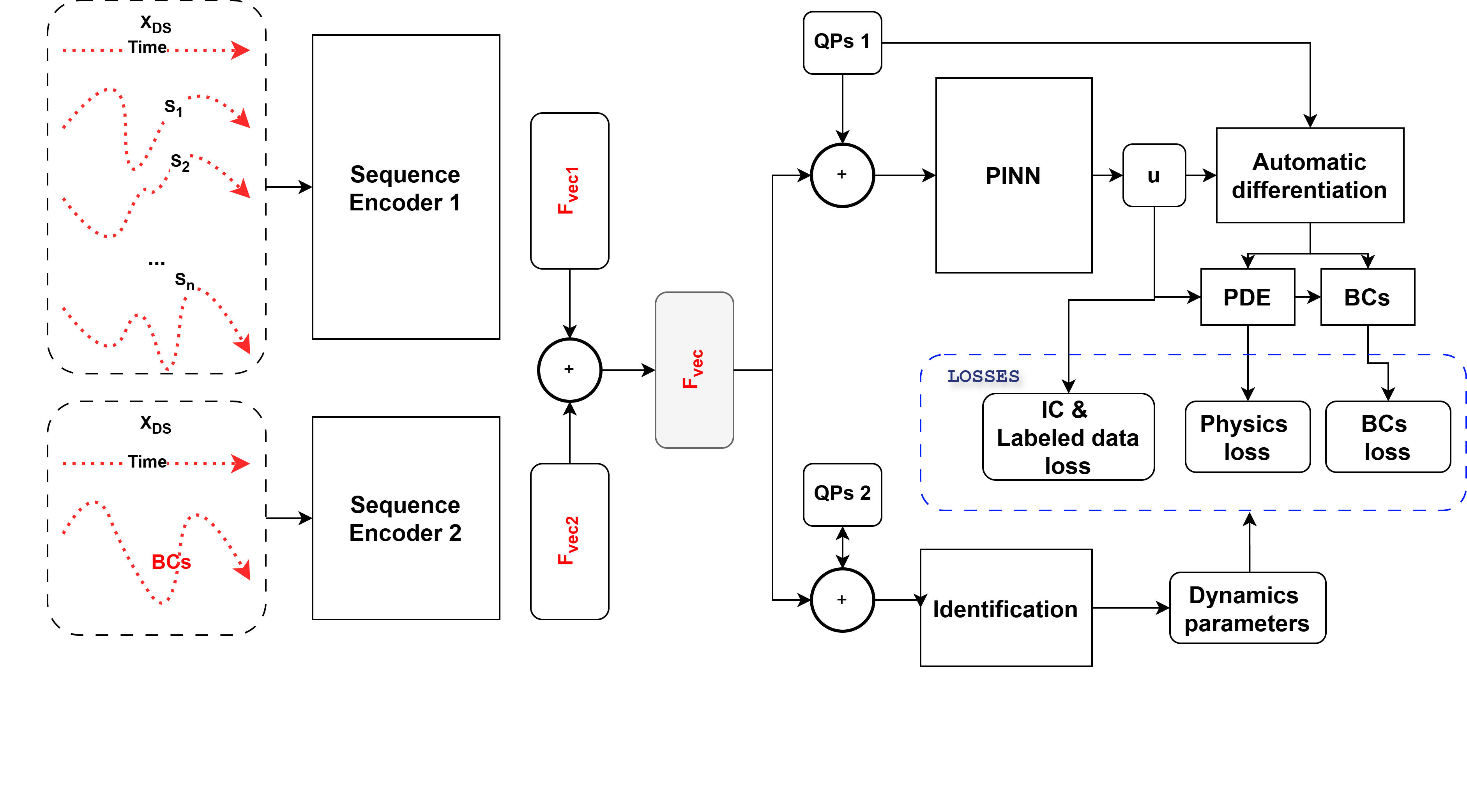}
    \caption{Overview of the PINN-SE model that combines two Sequences Encoders (Deep Sets or Set Transformer) with PINN and identification blocks, along with the training logic of the model. In this example architecture, we use two Sequence Encoders: the first to encode sensor data $s_i$, and the second to encode boundary conditions ($BCs$). Each SE generates a feature vector, which are then concatenated into a global feature vector $\mathbf{F}_{\text{vect}}$ and used by both the PINN and the identification blocks.}
    \label{PINN-SE-fig}
\end{figure}

Based on the idea of combining PINN and SINDy as presented in PINN-SR (Sec.~\ref{SINDy-PINN}), our approach follows a similar strategy by combining Deep Set and PINN. In our case, the Deep Set plays the role of encoding input data (such as sensors, BC/IC $\cdots$) and mapping it to the dynamic parameters or BCs/ICs. PINN will then use this information, along with the physics (PDE / ODE), to adapt to changes in dynamics as described in Fig~.\ref{PINN-SE-fig}. Additional blocks are used to predict the dynamics parameters or boundary conditions (BCs).
In this study, the inputs for the Deep sets are time series, time, and BC or positions (constant scalar), as shown in Fig.~\ref{SequenceEnc-fig}, and as given in Eq.~\ref{pinn-se-eq1}.

\begin{equation}\label{pinn-se-eq1}
\begin{aligned}
    \mathbf{SE}(time, & \, s_0, \dots s_i, \dots s_n,& \, BCs) = \mathbf{F_{vec}}
\end{aligned}
\end{equation}
where $\mathbf{F_{vec}}$ is the feature vector.\\
The time information is added to encode the sequence information, as the invariant neural network, by definition, does not have the sense of sequence. On the other hand, to make the model applicable in cases where $\Delta T$ is  variable (in most real world cases, the sensor acquisition frequency is variable), as demonstrated in the third  application (in Sec.~\ref{HT-App}).
In cases where the number of sensors or their positions are variable (Fig.~\ref{SequenceEnc-fig-DS}), Deep Set could be used as a separate encoder, as follows Eq.~\ref{pinn-se-eq2}:
\begin{equation}\label{pinn-se-eq2}
\begin{aligned}
    \mathbf{SE}_1(time, & \, space, \, s) &= \mathbf{F1_{vec}} \\
    \mathbf{SE}_2(time, & \, space, \, BCs) &= \mathbf{F2_{vec}}
\end{aligned}
\end{equation}

The first part, $\mathbf{SE}_1$, generates the feature vector, $\mathbf{F1_{vec}}$, for sensor data, and $\mathbf{SE}_2$, generates $\mathbf{F2_{vec}}$ for BCs. These two vectors can then be concatenated and used as input for the PINN prediction. It is also possible to use multiple encoders (Deep Sets) or to combine them with other architectures such as Convolutional Neural Network (CNN) (as in.~\cite{SETTRANS}).

The feature vector $\mathbf{F}_{\text{vec}}$ is then concatenated with the QPs, as shown in Fig.~\ref{PINN-SE-fig}. These QPs represent the spatial and temporal locations where we aim to predict the dynamics behavior. Specifically $\text{QPs}_1$ represent the points at which we want to predict the dynamical behavior using PINN, while $\text{QPs}_2$ are the points used to predict the dynamic parameters (in case where the dynamic parameters are time and space dependent). Note that the QPs are not necessarily the same as those provided as input to the Sequence Encoders. Instead, the core idea of using PINN is to leverage physical to extrapolate in space and time based on the encoded features. The concatenation operation between the feature vector and the QPs is defined in Eq.~\ref{pinn-se-eq3}.

\begin{equation}\label{pinn-se-eq3}
\begin{aligned}
    \mathbf{Y} &= \text{PINN}(\mathbf{X}) \\
    \mathbf{X} &= \text{concatenate}(\text{QPs},~\mathbf{F}_{\text{vec}}^r) \in \mathbb{R}^{\text{b} \times N \times (n_f + n_{qp})} \\
     \text{where} &\quad \mathbf{F}_{\text{vec}}^r = \text{repeat}(\mathbf{F}_{\text{vec}}) \in \mathbb{R}^{\text{b} \times N \times n_f}\\
     \text{and} & \quad \mathbf{F}_{\text{vec}} \in \mathbb{R}^{\text{b} \times 1 \times n_f} 
     , \quad \text{QPs} \in \mathbb{R}^{b \times N \times n_{qp}}.
\end{aligned}
\end{equation}

Where $\mathbf{Y}$ is the predicted dynamics state variables at the QPs, and $\mathbf{X}$ is the concatenation of the feature vector $\mathbf{F}_{\text{vec}}$ and the QPs. Since the QPs are used as input to the PINN, we can calculate the PDE residual of the model prediction using automatic differentiation, and in the same way, we can softly impose the BC/IC during training. It is also possible to use some labeled data to help the model converge faster.
In addition to the PINN block, we can also add other blocks to predict the dynamics parameters or the BCs (Identification model in Fig.~\ref{PINN-SE-fig}). In cases where the parameters are space- or time-dependent, we can concatenate \(\mathbf{F}_{\text{vec}}\) with \(\text{QPs}\) at points where we want to predict parameters, as shown in Fig.~\ref{PINN-SE-fig}. The idea behind adding these blocks is for inverse problem cases where the parameters are not known (similar to PINN-SR) or, by performing cross validation, we can validate the model's predictions.
The model could be trained using variable BCs and ICs. In these cases, we need to add this information to the Deep sets in order for the model to be able to encode it. In the next section, we test the model on different PDEs and ODEs, with different cases (variable BCs, ICs, parameters and geometries).

\section{Applications and results}\label{Applicationarai}
\subsection{Overview}\label{Appli_into}

In this section, we present three test case scenarios with different dynamics. In the first application, (Sec.~\ref{ODE-App}), we used the Rössler dynamics system defined as (Eq.~\ref{rosslerode}) : 
\begin{equation}\label{rosslerode} 
\begin{aligned} 
\frac{dx}{dt} &= - (y + z), \\ 
\frac{dy}{dt} &= x + a y, \\ 
\frac{dz}{dt} &= b + z (x - c). 
\end{aligned} 
\end{equation}

This ODE, though simple, describes the evolution of three state variables ($x$, $y$, and $z$) that can exhibit chaotic oscillations with respect to time $t$. This dynamic is parameterized with three parameters $a$, $b$, and $c$. For this study, we focus on generating data near to its chaotic behavior. The model is trained to predict the dynamics parameters using variable-length time series input and to extrapolate in time using physics. We will also study the possibility of using cross validation to provide some confidence level for the model's predictions.

The second test is an incompressible 2D Navier-Stokes problem, governed by the momentum balance and the conservation of mass as given in Eq.~\ref{NS-equation}:
\begin{equation}\label{NS-equation} 
\begin{aligned}
\frac{\partial \mathbf{u}}{\partial t} + (\mathbf{u} \cdot \nabla) \mathbf{u} &= -\frac{1}{\rho} \nabla p + \nu \nabla^2 \mathbf{u} , \\
\nabla \cdot \mathbf{u} &= 0.
\end{aligned}
\end{equation}
Where, $\mathbf{u}$ represents velocity through $x$ and $y$, and $p$ represents pressure, $\rho=1$ represents the fluid density, and $\nu=0.001$ is the kinematic viscosity. In this application, we study a laminar and Newtonian  flow around a cylinder with a parametric inlet velocity function. We investigate the model's ability to generalize to variable BC and show that only a few pressure sensors are needed to predict the inlet velocity function and provide the 2D velocity and pressure fields throughout the entire domain, using physics.\\
In the third application we validated the model using acquired experimental data from the heating of thermoplastic composite material plates in an infrared oven. Two contact-less sensors, placed at the top and bottom of the plate were used to capture the temperature during the heating process. In this study, we used two types of composite materials with different thermal properties, BCs, and thicknesses. We also examined the model's capability to detect changes in heat power percentage and adapt to them. The heat transfer problem in this case can be simplified to 1D (Eq.~\ref{HT-eq}):
\begin{equation}\label{HT-eq} 
\begin{aligned}
\frac{\partial T}{\partial t} &= \alpha \frac{\partial^2 T}{\partial x^2}, \quad 0 < x < L, \quad t > 0, \\
-k \frac{\partial T}{\partial x} \Big|_{x=0} &= q_0 - h (T(0,t) - T_{\infty}), \\
-k \frac{\partial T}{\partial x} \Big|_{x=L} &= q_L - h (T(L,t) - T_{\infty}), \\
T(x,0) &= T_0(x).
\end{aligned}
\end{equation}
Where $T$ is the temperature, $\alpha$ is the diffusivity, $h$ is the heat exchange coefficient, $k$ is the conductivity, $L$ is the thickness of the plate, and $q_0$ and $q_L$ represent the imposed heat flux at the top and bottom of the plate, respectively.

\subsection{Application 1 : Rössler system, model generalization, robustness to noise and adaptability}\label{ODE-App}
\subsubsection{Data generation}\label{rossler_data_gen}
\begin{figure}[!h]
    \centering
    \includegraphics[width=0.7\textwidth]{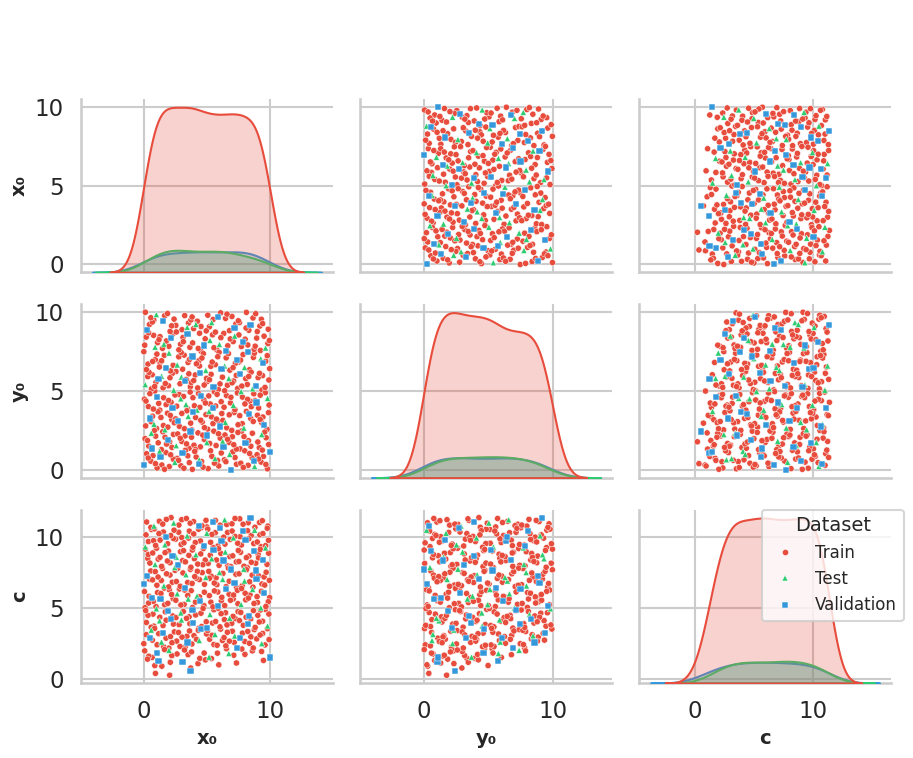}
    \caption{Distribution of initial conditions \(x_0\) and \(y_0\), and the Rössler dynamics parameter \(c\) for training, validation, and testing. During the simulation, we eliminated divergence cases, which can be seen for examples in the distribution of \(y_0\) and \(c\).}
    \label{Distribution-rossler}
\end{figure}

\begin{figure}[!h]
    \centering
    \includegraphics[width=0.7\textwidth]{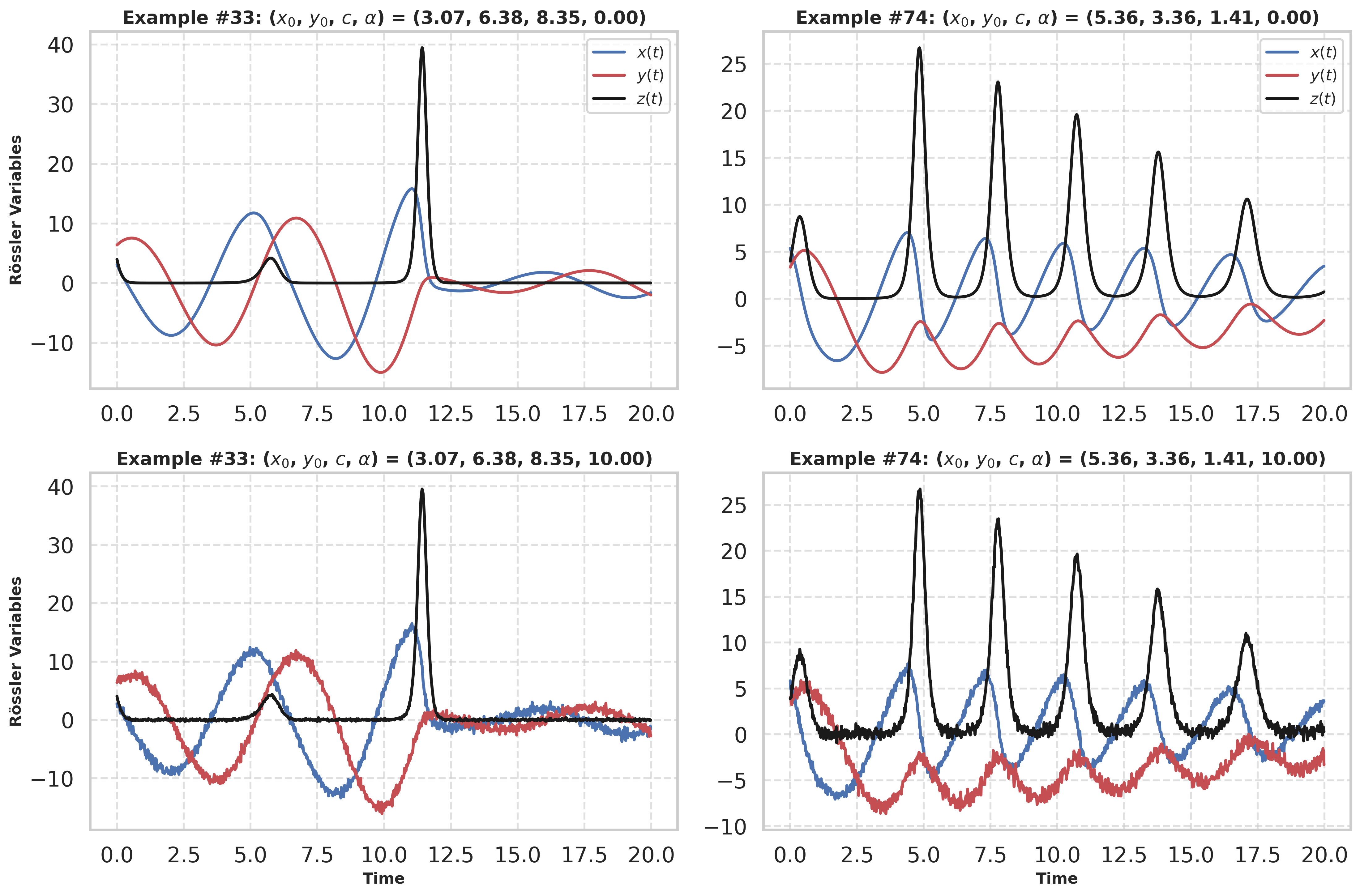}
    \caption{Example of the Rössler system with different parameters \(c\), initial conditions \(x_0\), \(y_0\) and noise level $\alpha$ illustrating the system's dynamic behavior.}
    \label{rossler_example}
\end{figure}

The model is trained using synthetic data generated with the LSODA~\cite{LSODA} (Livermore Solver for Ordinary Differential Equations) ODE solver, implemented in Python via the SciPy~\cite{SCIPY} Python package. In this application, we used 3 variations for data generation: IC ($x_0$, $y_0$) and the parameter $c$ of the Rössler dynamics, while $a = b = 0.2$ are fixed near the chaotic behavior of Rössler dynamics, and $z_0 = 4$. \\
The distribution of these 3 variations is given in Fig.~\ref{Distribution-rossler}. For these variations, we generated 453 datasets using Sobol~\cite{SOBOL} sequences to ensure a good distribution of all possible combinations. For each dataset (Fig.~\ref{rossler_example}), the simulation time $T$ is 20 seconds, with a sampling rate of $f=1/\Delta T$, where $\Delta T = 0.01~s$, which means that for each dataset, we have 3 sequences of variables $x(t)$, $y(t)$, and $z(t)$, with 2000 time points each.\\

We then introduced 10\% of noise to the synthetic dynamics. We added Gaussian white noise (Fig~\ref{rossler_example}), with its intensity determined by the root mean square (RMS) of the exact solution as defined in Eq.~\ref{noiseadded}:\\
\begin{equation}\label{noiseadded}
\begin{aligned}
    I_{\text{noisy}}^j &= I_{\text{exact}}^j + \eta \cdot \sigma, \quad \eta \sim \mathcal{N}(0,1), \quad j\in \{x(t),~y(t),~z(t)\}, \\
    \sigma &= \alpha \cdot \frac{\sqrt{\sum_{i=1}^{N} I_{\text{exact}}^{j}(t_i)^2}}{N}, \quad \alpha \equiv \text{ratio} \in [0\%,~10\%], \quad j\in \{x(t),~y(t),~z(t)\}
\end{aligned}
\end{equation}

\subsubsection{Model architecture and training}\label{Model architecture and training}
\begin{figure}[!h]
    \centering
    \includegraphics[width=0.7\textwidth]{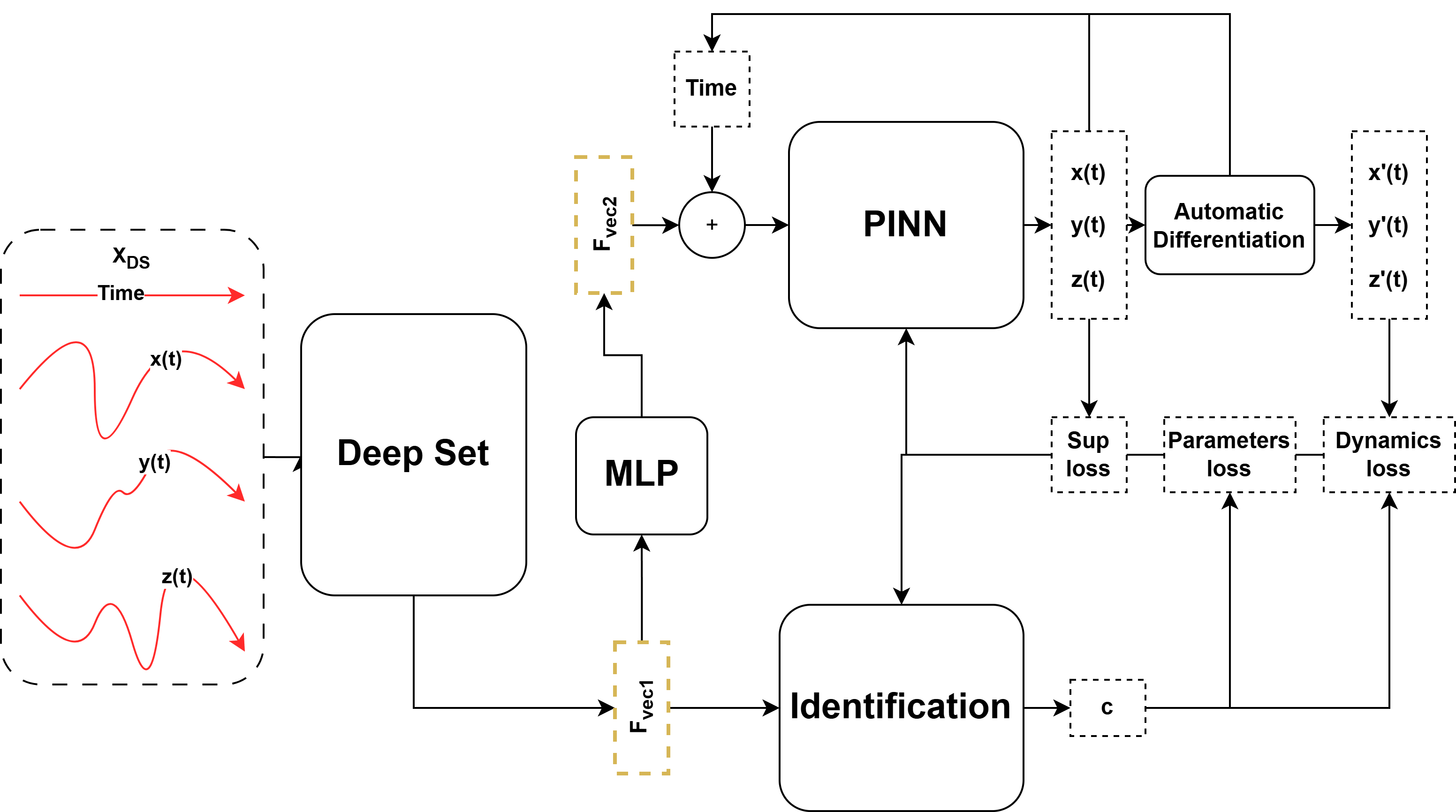}
    \caption{The PINN-SE model used for the first application (Rössler dynamics) operates as follows: the Deep set takes as input the $time$ and state variables $x$, $y$, and $z$, given as  $X_{DS} \in \mathbb{R}^{\text{b} \times N \times 4}$, and predicts feature vector $\mathbf{F}_{\text{vect}} \in \mathbb{R}^{b \times 1 \times d_{f.v}}$. This feature vector is then used directly by the identification blocks (MLPs) to predict the parameter $c$. The feature vector is also concatenated with the time snapshots for which we want to predict the state variables $x$, $y$, and $z$ using the PINN blocks. The dynamics loss (Eq.~\ref{loss_rossler}) is calculated using automatic differentiation.}
    \label{rossler_model_archi}
\end{figure}

\begin{figure}[!h]
    \centering
    \includegraphics[width=.75\textwidth]{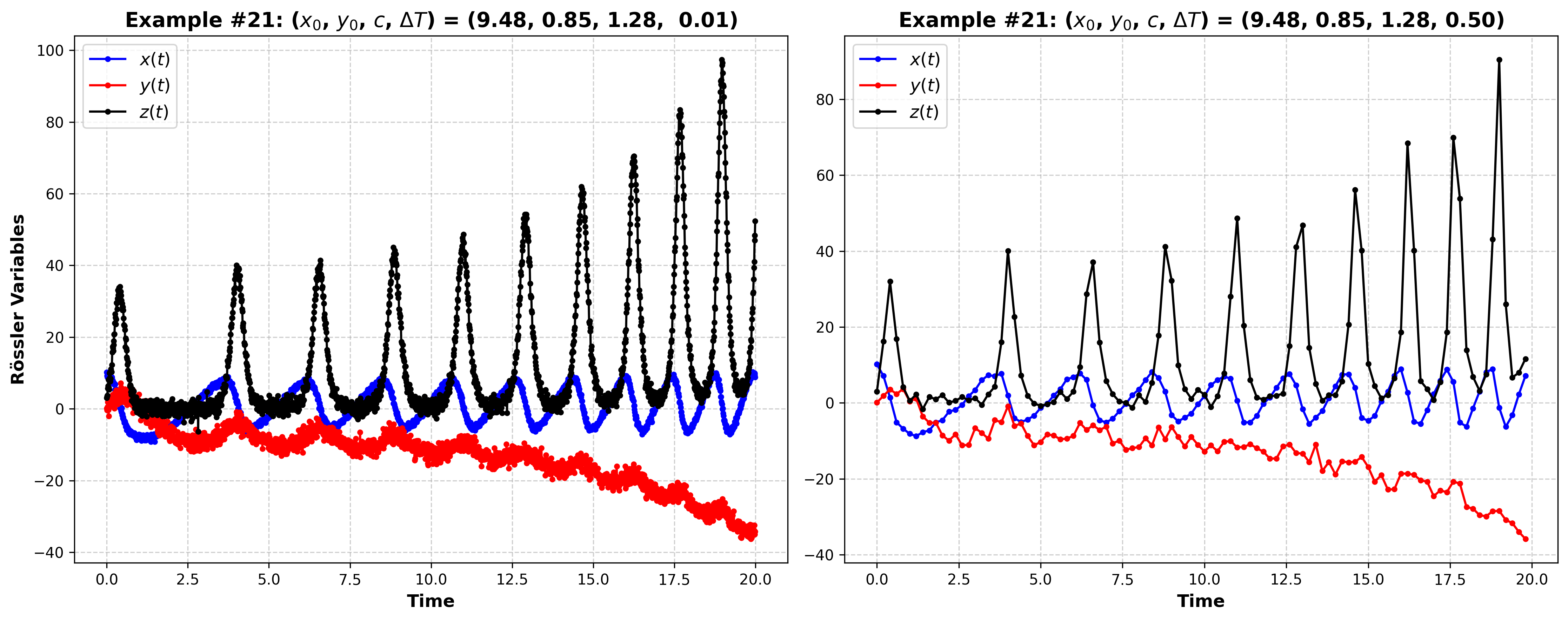}
    \caption{An example of the Rössler dynamic with variable $\Delta T$ used for model training, both as Deep set input and for the supervised loss : on the left with $\Delta T = 0.01$, and on the right with $\Delta T = 0.5$.}
    \label{rossle_exampl_var_dt}
\end{figure}

The model is trained using the expanding window method (Example in Fig.~\ref{rossle_exampl_var_dt}). This approach leverages the advantages of Deep sets, that can encode variable sequence lengths and map them to dynamic behaviors. This means the model is trained to integrate new time snapshot data to update its predictions. In this application, the model is trained using 362 datasets selected randomly (using Sobol sampling), the rest is used for validation and testing. The model architecture is given in Fig.~\ref{rossler_model_archi}. The model input, $X_{DS}$, represents time and corresponding state variable snapshots. We use a time step of $\Delta T_{DS} = 0.5~s$ for the Deep sets input. 
The input tensor is $X_{DS} \in \mathbb{R}^{b \times N \times d}$, where $b$ is the batch size, $N \in [10, 40]$ is the number of snapshot, and $d = 4$ is the input dimension (time, $x$, $y$, and $z$). This input is projected to a higher-dimensional representation $\Phi \in \mathbb{R}^{b \times N \times d_{f.v}}$, \textcolor{red}{where $d_{f.v}$ is the dimension of the feature vector}. 
We then apply an aggregation function to compute the feature vector $\mathbf{F}_{\text{vect}} \in \mathbb{R}^{b \times 1 \times d_{f.v}}$. The choice of aggregation function is crucial and must be made carefully to ensure generalization. In this application, we used the following aggregation Eq.~\ref{agg-appli1}:

\begin{equation}\label{agg-appli1}
\left\| .\right\|_2^2 \equiv\sum_{i=1}^{N} \frac{\Phi_{i}^2}{N} \quad,  \textcolor{red}{\text{where}} \quad \textcolor{red}{\Phi = \left[ \Phi_{1} ... \Phi_{i}... \Phi_{N} \right]} \quad \textcolor{red}{\Phi_{i} \in \mathbb{R}^{b \times 1 \times d_{f.v}}}
\end{equation}

The characteristic vector is then concatenated with time (query points), as shown in Fig.~\ref{rossler_model_archi}, where the goal is to predict the system dynamics. This concatenation is defined in Eq.~\ref{conca_rossler}:

\begin{equation}\label{conca_rossler}
\begin{aligned}
\textbf{X}_{pinn}& = \textrm{concatenate}(\mathbf{F}_{\text{vect}}, ~ t)\\
\text{where} & ~~\mathrm{time} \in \mathbb{R}^{b\times N_{q.p}\times 1} \\
\text{and} & ~~\mathbf{F}_{\text{vect}}\in \mathbb{R}^{b\times 1\times d_{f.v}}\\
\text{and} & ~~ \textrm{concatenate}(\mathbf{F}_{\text{vect}}, ~ t) = \begin{bmatrix} \mathbf{F}_{\text{vect}}, ~ t \end{bmatrix} \in \mathbb{R}^{b\times N_{q.p} \times (d_{f.v} + 1)}
\end{aligned}
\end{equation}
, where $N_{q.p}$ represents the query points (snapshots that are different from those provided as input to the Deep set) at which we aim to predict the Rössler dynamics.

The model is trained to predict both the system dynamics $x(t)$, $y(t)$, and $z(t)$, as well as the parameter $c$. The motivation behind this approach is to enable cross validation of the model's predictions. The loss function is defined in Eq.~\ref{loss_rossler}:\\

\begin{equation}\label{loss_rossler}
\begin{aligned}
\mathcal{L}_{\text{total}} &= \mathcal{L}_{\text{sup}} + \lambda \mathcal{L}_{\text{ODE}} + \beta \mathcal{L}_{\text{P}}, \\
\mathcal{L}_{\text{sup}} &= \|I(t) - \widetilde{I(t)}\|_1, \\
\mathcal{L}_{\text{ODE}} &= \|\dot{\widetilde{x}} + \widetilde{y} + \widetilde{z}\|_1 + \|\dot{\widetilde{y}} - \widetilde{x} - a \widetilde{y}\|_1 + \|\dot{\widetilde{z}} - b - \widetilde{z}(\widetilde{x} - c)\|_1, \\
\mathcal{L}_{\text{P}} &= \|P\|_2^2
\end{aligned}\\
\end{equation}
Where $\mathcal{L}_{\text{sup}}$ is the supervised loss, and $I(t) = [x(t),~y(t),~z(t)]$ represents the labeled noisy points used for supervision loss. We employed sparse input sampling with $\Delta T_{\text{sup}} = 0.5$ for the supervised loss. The model prediction, $\widetilde{I(t)}$, is computed as shown in Eq.~\ref{pinn-se-app1_sup}:\\

\begin{equation}\label{pinn-se-app1_sup}
\begin{aligned}
    \widetilde{I(t)} &= \text{PINN}(\mathbf{X}) \\
    \mathbf{X} &= \text{concatenate}(t,~\mathbf{F}_{\text{vec}}) \in \mathbb{R}^{\text{b} \times 40 \times (n_f + 1)} \\
    \text{where} \quad \mathbf{F}_{\text{vec}} &\in \mathbb{R}^{\text{b} \times 1 \times n_f}
\end{aligned}
\end{equation}

The second component $\mathcal{L}_{\text{ODE}}$ is the dynamics loss, where $c$ represents the true dynamics parameter. The dynamics loss is used to predict dynamic behavior at time instants where no supervised data is available. The third component $\mathcal{L}_{\text{P}}$ corresponds to the loss for predicting the parameter $c$.

Using validation datasets, we identified an optimal model architecture by manually tuning the model’s hyperparameters, including the aggregation function, standard MLP hyperparameters, and the loss function weights $\lambda$ and $\beta$. The final selected architecture is presented in Tab.~\ref{Table_app_1_archi}.\\

\begin{table}[!hpt]
\caption{The architecture details of the model, including the Deep Set, Pooling, Post-Pooling, PINN, and Identification blocks, are provided. The table specifies the width, number of layers, activation functions, and additional comments for each component.}\label{Table_app_1_archi}
\begin{tabular}{|l|lll|l|}
\hline
\textbf{}               & \multicolumn{1}{l|}{\textbf{Width}} & \multicolumn{1}{l|}{\textbf{Layers}} & \textbf{Activation}              & \textbf{Comments}        \\ \hline
\textbf{Deep set}       & \multicolumn{1}{l|}{80}             & \multicolumn{1}{l|}{4}               & $\text{gelu}(X)$                    & -                        \\ \hline
\textbf{Pooling}        & \multicolumn{3}{c|}{-}                                                                                        & $\left\|. \right\|_2^2$    \\ \hline
\textbf{Post-Pooling}   & \multicolumn{1}{l|}{80}             & \multicolumn{1}{l|}{3}               & $\text{gelu}(X)$                   & -                        \\ \hline
\textbf{PINN}           & \multicolumn{1}{l|}{80}             & \multicolumn{1}{l|}{4}               & $\text{gelu}(X)\times(1+\sin(3.14X))$ & for each state variables \\ \hline
\textbf{Identification} & \multicolumn{1}{l|}{80}             & \multicolumn{1}{l|}{3}               & $\text{gelu}(X)$                    & -                        \\ \hline
\end{tabular}

\end{table}
The model training process can be summarized in Algo.~\ref{alog_app1}.

\begin{algorithm}[!hpt]
\caption{The training strategy for the Rössler dynamics is presented, including the evolution of the learning rate during training and the parameters of the loss function, with an Epoch Loop and dynamic $\lambda$ adjustment.}\label{alog_app1}
\begin{algorithmic}[1]
\State \textbf{Step 1: Initialize Training}
\State Set learning rate: \(lr = 6 \times 10^{-3}\)
\State Set loss weights: \(\lambda = 0\), \(\beta = 1\)
\State Set Trainable variables: all PINN-SE blocks, including Deep Set

\For{epoch = 1 to max\_epochs}
    \State \textbf{Train Loop}

    \If{$lr > 1 \times 10^{-3}$}
        \State \(lr \gets lr \times 0.999\) per epoch
    \Else
        \State \(lr \gets lr \times 0.9999\)
    \EndIf

    \If{Validation error of parameters $<$ threshold}
        \State Deactivate Deep Set training 
    \EndIf

    \If{$\mathcal{L}_{\text{SUP}} <$ threshold}
        \State Increase \(\lambda \gets \lambda + 0.0001\) per epoch
    \Else
        \State Decrease \(\lambda \gets \lambda \times 0.99\) per epoch
    \EndIf
\EndFor
\end{algorithmic}
\end{algorithm}

\subsubsection{Results and analysis}
\begin{figure}[!h]
    \centering
    \includegraphics[width=1.\textwidth]{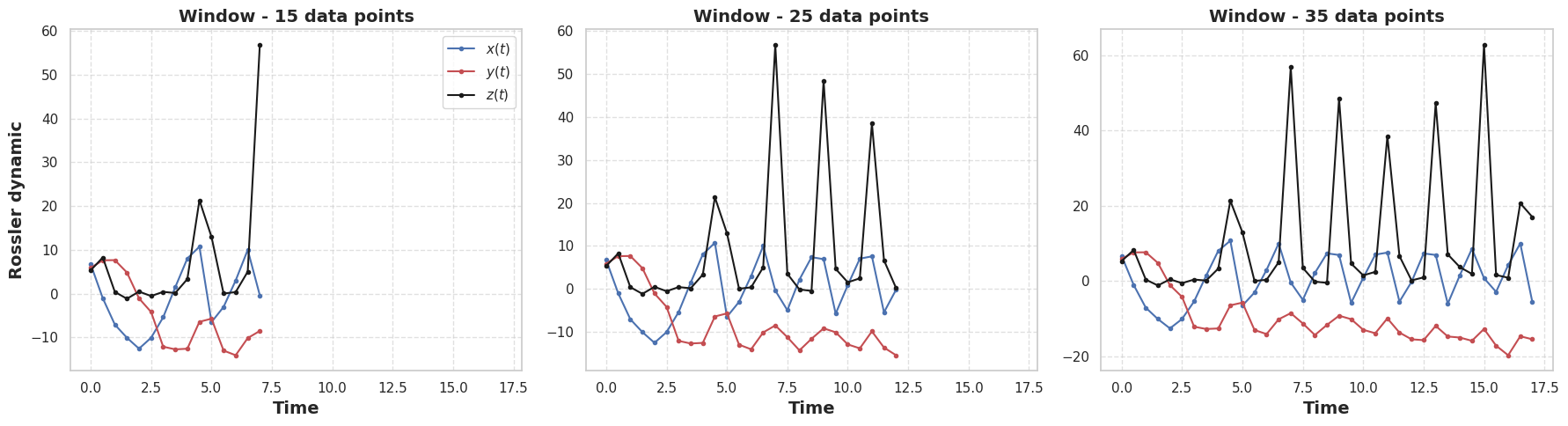}
    \caption{Deep Set input as sequences: Three examples of variable length windows (15, 25 and 35 respectively from left to right) with a time steps of $\Delta T_{DS} = 0.5~\text{s}$.}
    \label{expanding wind}
\end{figure}

The model's performance is assessed based on its ability to predict the dynamic parameter $c$ and interpolate and extrapolate over time. We also examine whether the model can use the dynamic information and the identification blocks for cross validation of its predictions.\\  
As described in Sec.~\ref{rossler_data_gen}, we used 362 datasets for training, 45 for validation and model selection, and 46 for testing its performance. As outlined in Sec.~\ref{Model architecture and training}, the Deep set takes as input $\text{time}$, $x(t)$, $y(t)$, and $z(t)$ with variable sequence lengths. In this study, we evaluated the model using five different sequence lengths: [15, 20, 25, 30, 35], as shown in Fig.~\ref{expanding wind}.

\begin{figure}[!h]
    \centering
    \includegraphics[width=0.7\textwidth]{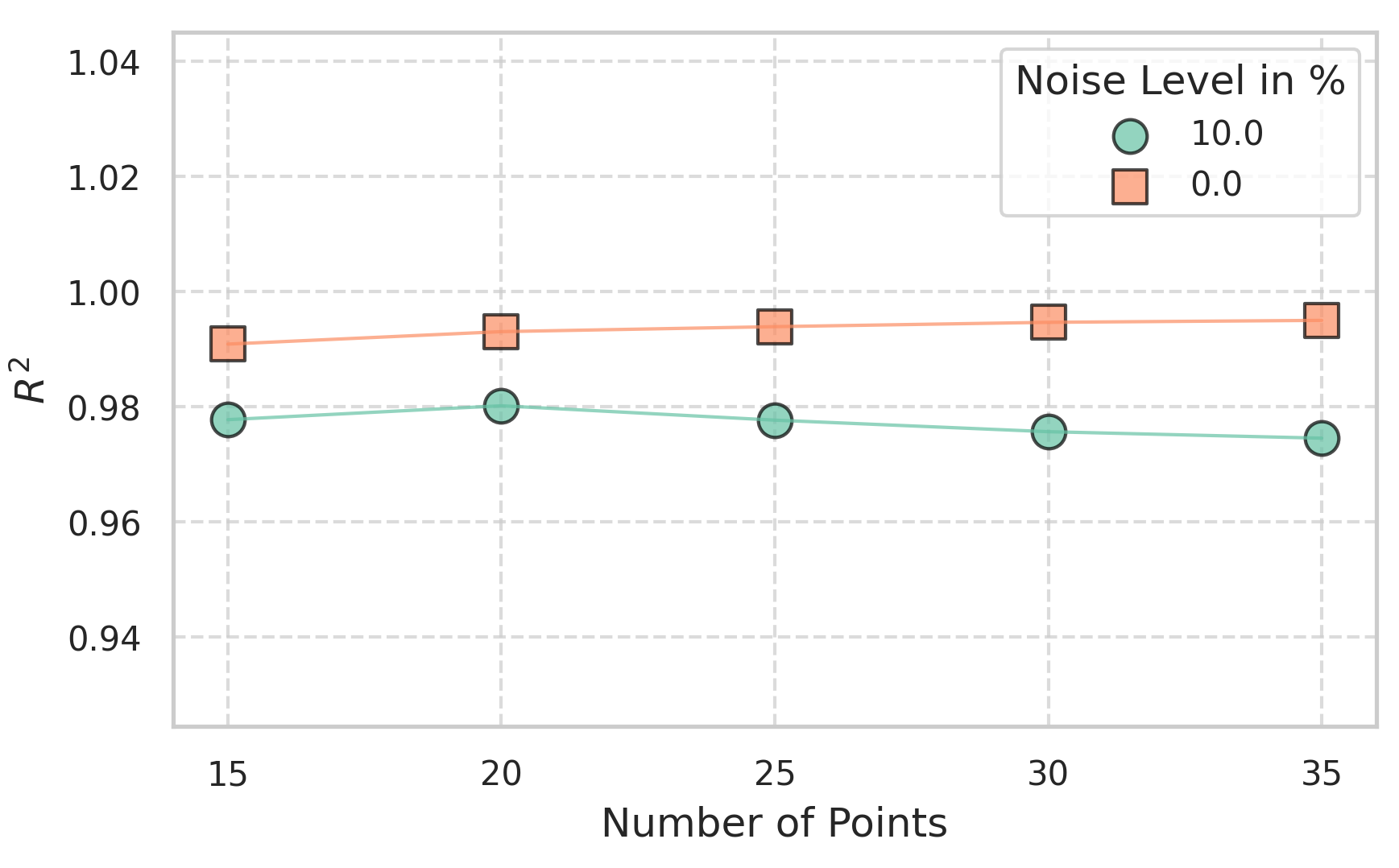}
    \caption{Model accuracy using the $R^2$ metric for predicting the Rössler dynamics parameter $c$ with respect to variable input window sizes ranging from 15 to 35, for both the example with 0\% noise and the one with 10\% added noise.}
    \label{accu_c_wrt_noise}
\end{figure}

We tested the model using two examples: one with 0\% noise and another with 10\% added noise (Eq.~\ref{noiseadded}). We also study the Deep set's ability to encode variable time series lengths to improve its predictions. In Fig.~\ref{accu_c_wrt_noise}, we evaluate the model's ability to capture the dynamic parameter $c$. We used the $R^2$ metric~\cite{chicco2021coefficient} to assess the accuracy of the model's predictions.

\begin{figure}[!h]
    \centering
    \includegraphics[width=0.7\textwidth]{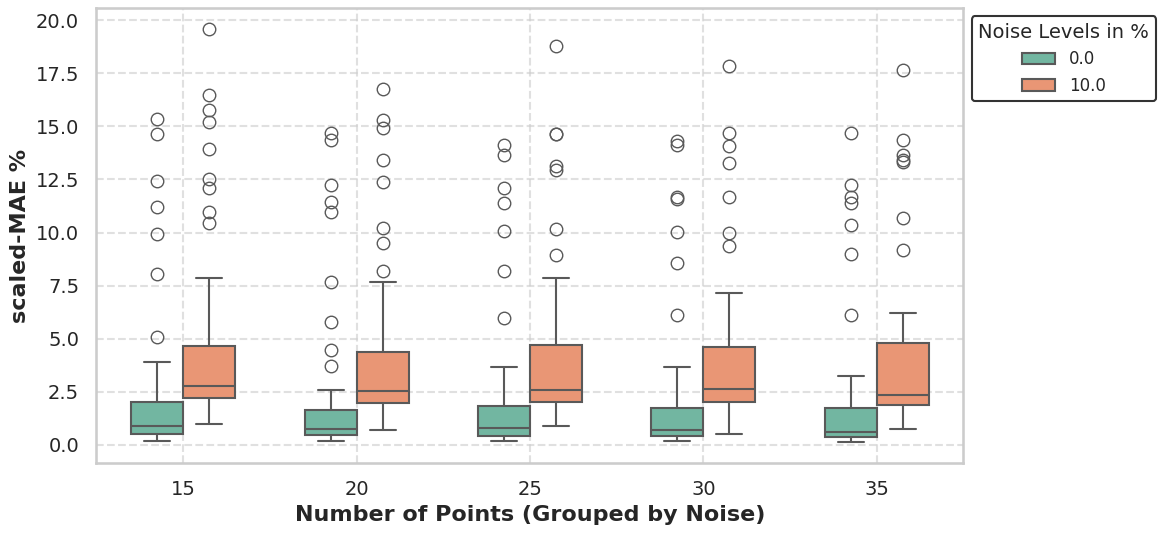}
    \caption{Model accuracy for predicting the state variable $x(t)$, where $t \in [0, 20]$, of the Rössler dynamics, assessed using scaled-\textbf{MAE}, with varying Deep Set input lengths for both noisy and non-noisy data. On the x-axis, we have the Deep Set sequence input lenghts (window size), and on the y-axis we have the scaled-\textbf{MAE}. The model accuracy is good; nevertheless, for some examples (outliers noted with open circles, 10\% of the total test dataset), the model's prediction is not very accurate.}
    \label{sMAE_x}
\end{figure}

\begin{figure}[!h]
    \centering
    \includegraphics[width=0.7\textwidth]{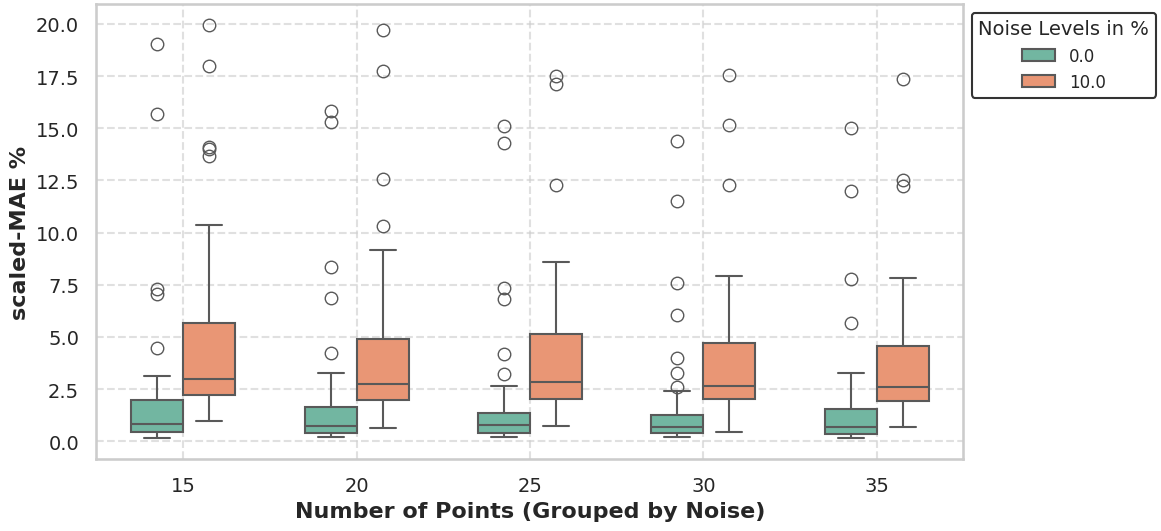}
    \caption{Model accuracy for predicting the state variable $y(t)$, where $t \in [0, 20]$, of the Rössler dynamics, assessed using scaled-\textbf{MAE}, with varying Deep Set input lengths for both noisy and non-noisy data. On the x-axis, we have the Deep Set sequence input lenghts (window size), and on the y-axis we have the scaled-\textbf{MAE}. The model accuracy is good; nevertheless, for some examples (outliers noted with open circles, 10\% of the total test dataset), the model's prediction is not very accurate.}
    \label{sMAE_y}
\end{figure}

\begin{figure}[!h]
    \centering
    \includegraphics[width=0.7\textwidth]{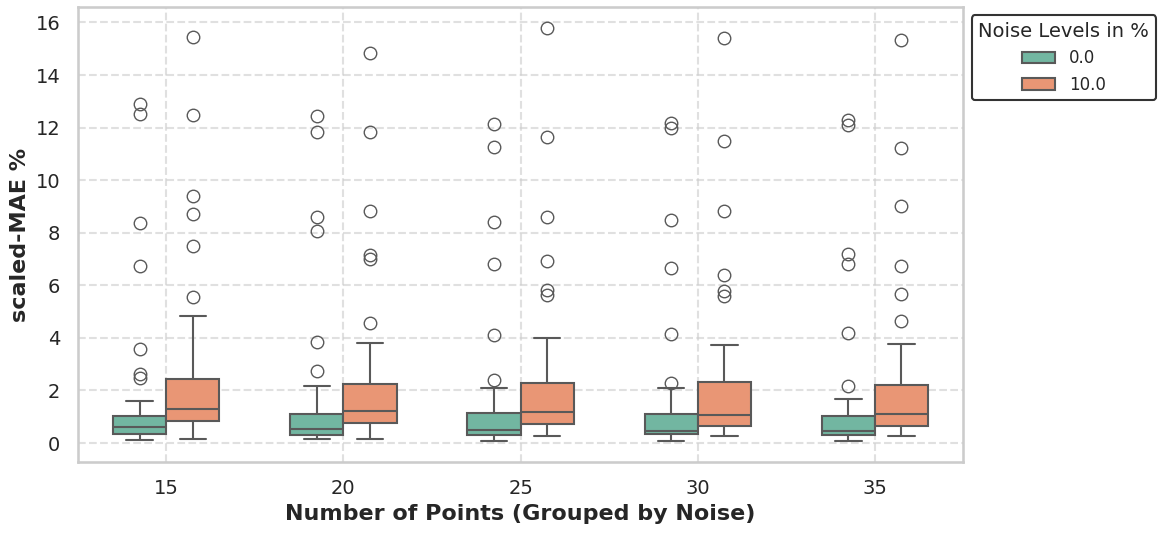}
    \caption{Model accuracy for predicting the state variable $z(t)$, where $t \in [0, 20]$, of the Rössler dynamics, assessed using scaled-\textbf{MAE}, with varying Deep Set input lengths for both noisy and non-noisy data. On the x-axis, we have the Deep Set sequence input lenghts (window size), and on the y-axis we have the scaled-\textbf{MAE}. The model accuracy is good; nevertheless, for some examples (outliers noted with open circles, 10\% of the total test dataset), the model's prediction is not very accurate.}
    \label{sMAE_z}
\end{figure}

\begin{figure}[h!]
    \centering
    \begin{minipage}{0.9\textwidth}
        \centering
        \includegraphics[width=0.8\linewidth]{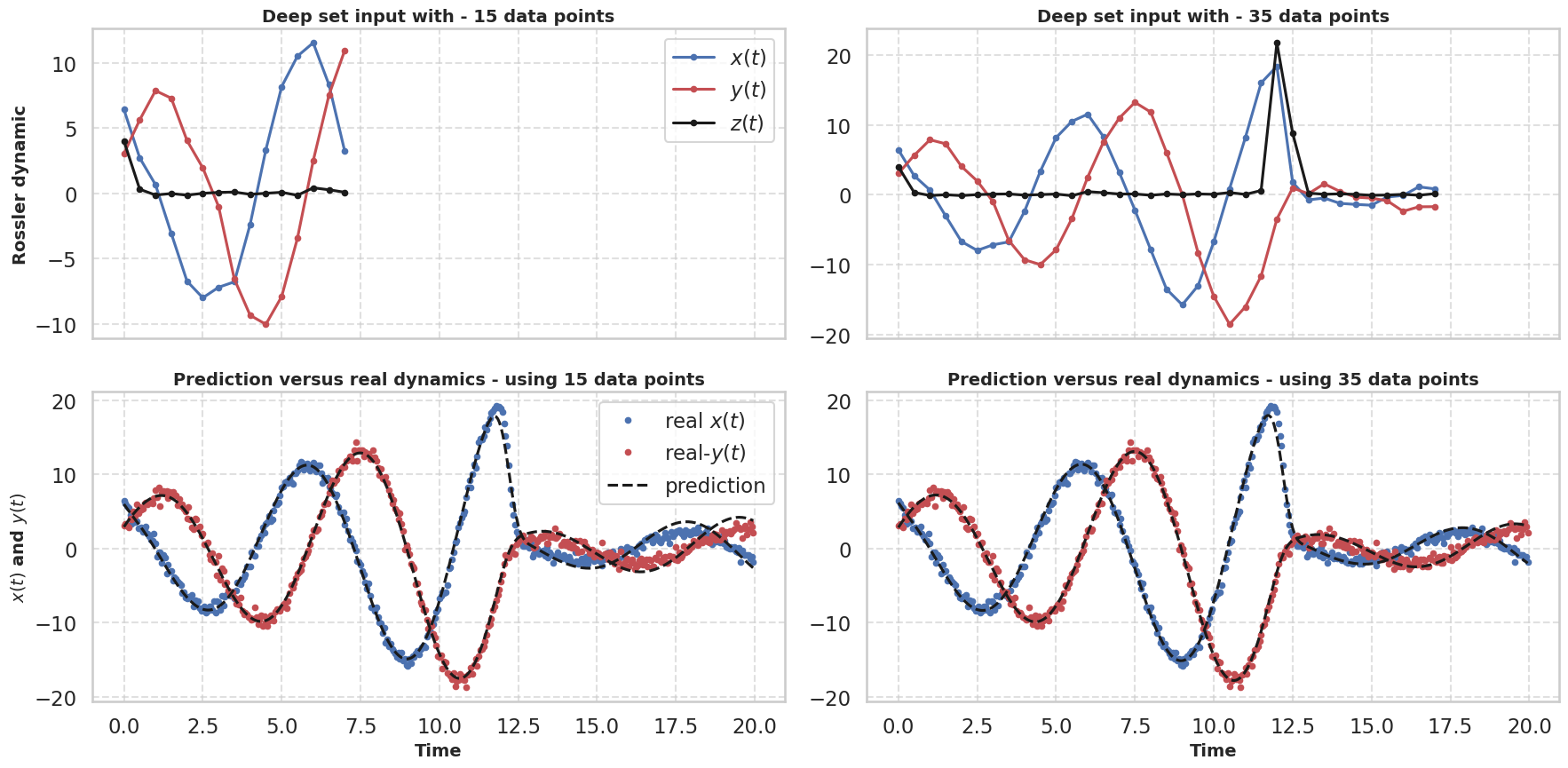} 
        \caption{Example of a good prediction by the model. At the top, we show the Deep set inputs with variable data points (variable lengths), and at the bottom, we present the corresponding model outputs for each number of points given as input, alongside the real dynamics (with added noise).}\label{MAE_low_}
    \end{minipage}\hfill
    \begin{minipage}{0.9\textwidth}
        \centering
        \includegraphics[width=0.8\linewidth]{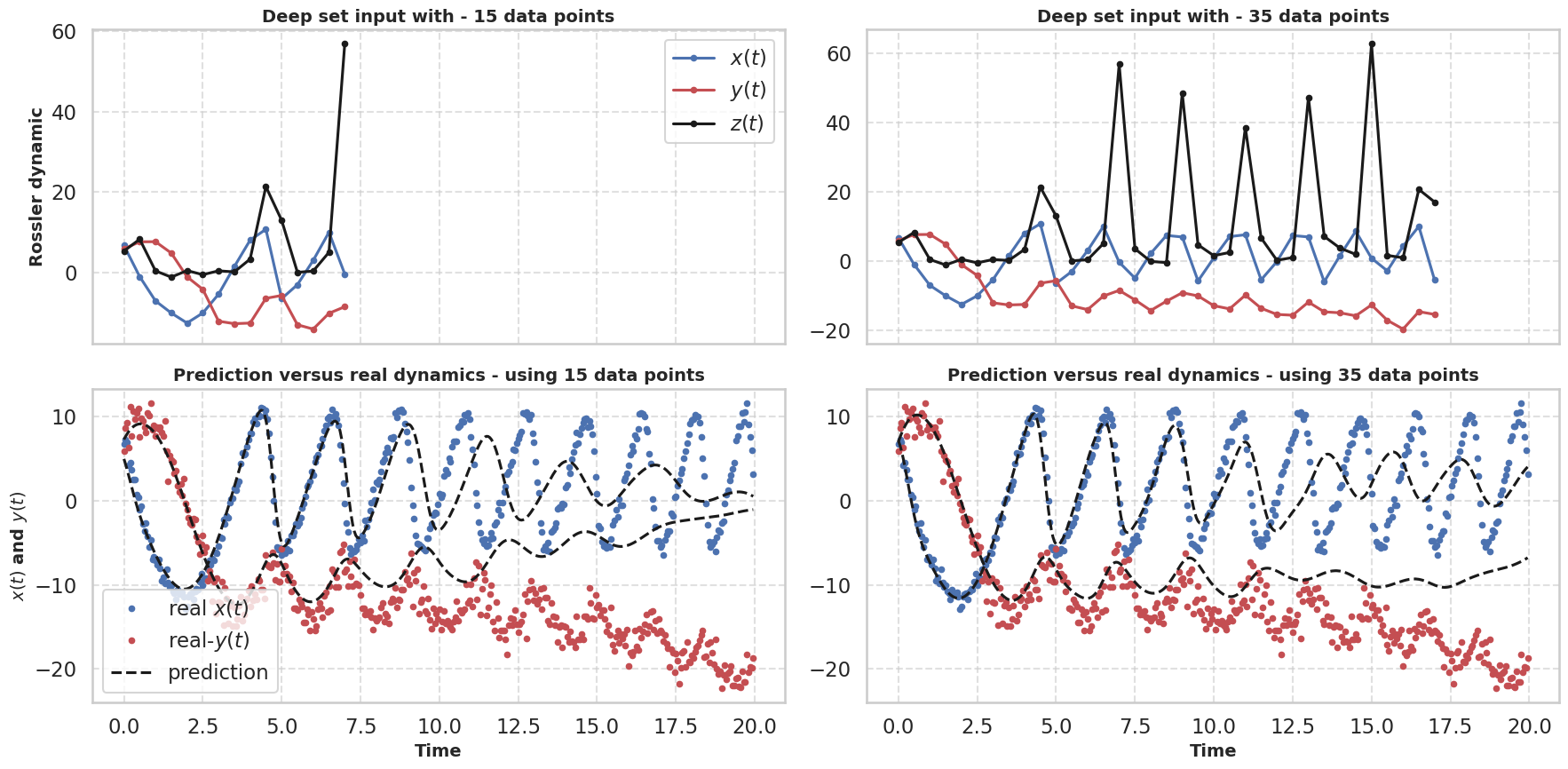} 
        \caption{Example of an outlier dataset where the model's prediction is not very accurate. At the top, we show the Deep Set inputs with variable data points (variable lengths), and at the bottom, we present the corresponding model outputs for each number of points given as input, alongside the real dynamics (with added noise).}\label{MAE_high_}
    \end{minipage}
\end{figure}

As shown in Fig.~\ref{accu_c_wrt_noise}, the model can easily capture the dynamic parameter $c$ with few data points. It also demonstrates strong stability with respect to noise. We also tested the model's accuracy to predict the dynamics behavior ($x$, $y$, and $z$). The feature vector used to predict the parameter $c$ is also utilized by the PINN to predict the dynamics. \\
As stated earlier, we used a time step $\Delta T_{\text{train}} = 0.5$ for the supervised loss, and the dynamics loss was applied to the rest of the time domain. To assess the model's ability to leverage physics for improved predictions, we tested it using $\Delta T_{\text{test}} = 0.05$. The error was computed over the entire time domain $t \in [0,~20]$. In this case, we used the Mean Absolute Error (\textbf{MAE}), specifically a scaled-\textbf{MAE}, as a metric to assess the model's prediction accuracy (Eq.~\ref{metrics_app1}, where $n$ is the number of time points, for $t \in [0,~20]$, and $\Delta T_{\text{test}} = 0.05$, with $n=400$).

\begin{equation}\label{metrics_app1}
\begin{aligned}
    \text{MAE} &= \frac{1}{n} \sum_{i=1}^{n} |I_i - \hat{I}_i| \quad \text{where} \quad I_i \in \{x(t_i), y(t_i), z(t_i)\} \\
    \text{Scaled-MAE} &= 100 \cdot \frac{\text{MAE}}{\max(I) - \min(I)} 
\end{aligned}
\end{equation}
As the Mean Absolute Percentage Error (\textbf{MAPE}) metric could not be used due to the presence of outliers~\cite{FORCASTBP} in $z(t)$, and since scaled-\textbf{MAPE} (\textbf{sMAPE}) is also not appropriate because $z(t)$ is null~\cite{FORCASTBP} (zero in the denominator) for most of the time until it increases significantly at a certain point (Fig.~\ref{rossler_example}), we instead present the results using scaled-\textbf{MAE} in Figs.~\ref{sMAE_x}, \ref{sMAE_y}, and \ref{sMAE_z} for $x(t)$, $y(t)$, and $z(t)$, respectively, for all test datasets. In Fig.~\ref{MAE_low_}, we present two examples where the model's prediction is accurate (scaled-\textbf{MAE} less than 10\%), as well as an example illustrating the outliers of the scaled-\textbf{MAE} (Fig.~\ref{MAE_high_}).

These results show that the model has very good accuracy and a good stability with respect to noise presence, as the error does not increase significantly. Additionally, we observe the model's ability to encode variable sequence lengths to improve its predictions (Fig.~\ref{sMAE_y}).

\begin{figure}[!h]
    \centering
    \includegraphics[width=0.7\textwidth]{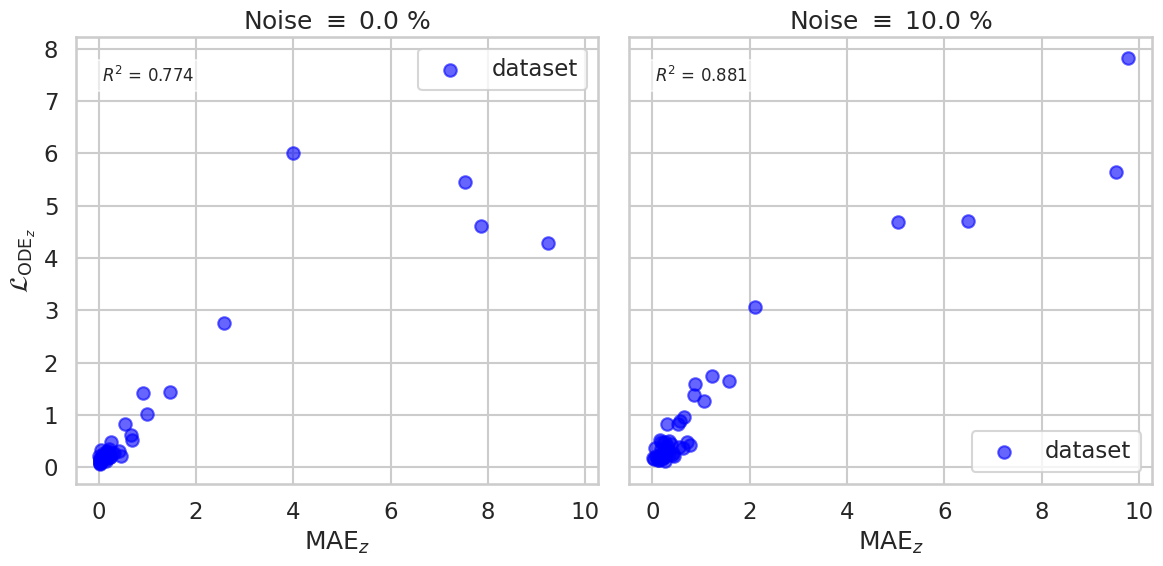}
    \caption{Correlation results from the cross validation between the Identification and PINN prediction for all the test datasets (points), where the x-axis represents the MAE of the prediction of $z(t)$, and the y-axis represents the physics residual $\mathcal{L}_{\text{ODE}_z}$ (using 25 data points as input to the Deep set).}
    \label{r2ode_mae}
\end{figure}

We also tested the ability to use cross validation between the identification block and the PINN to provide a confidence level for the model's prediction. To do so, we calculated the residual of the third equation of the Rössler dynamics (eq.~\ref{rosslerode}, $\mathcal{L}_{\text{ODE}}$), by computing $\dot{z}$ using automatic differentiation and the Rössler dynamics corresponding to $b + \widetilde{z}(\widetilde{x} - \widetilde{c})$, where $\widetilde{c}$ is the prediction of the identification model. This was used to calculate the ODE loss $\mathcal{L}_{\text{ODE}_z}$ and $\mathbf{MAE_z}$. In Fig.~\ref{r2ode_mae}, we present the correlation between $\mathcal{L}_{\text{ODE}_z} = \left\| \dot{z} - b + \widetilde{z}(\widetilde{x} - \widetilde{c})\right\|_1$ and the $\mathbf{MAE_z}$ loss for the state variable $z$ across all the test datasets.

\begin{figure}[H]
    \centering
    \includegraphics[width=0.7\textwidth]{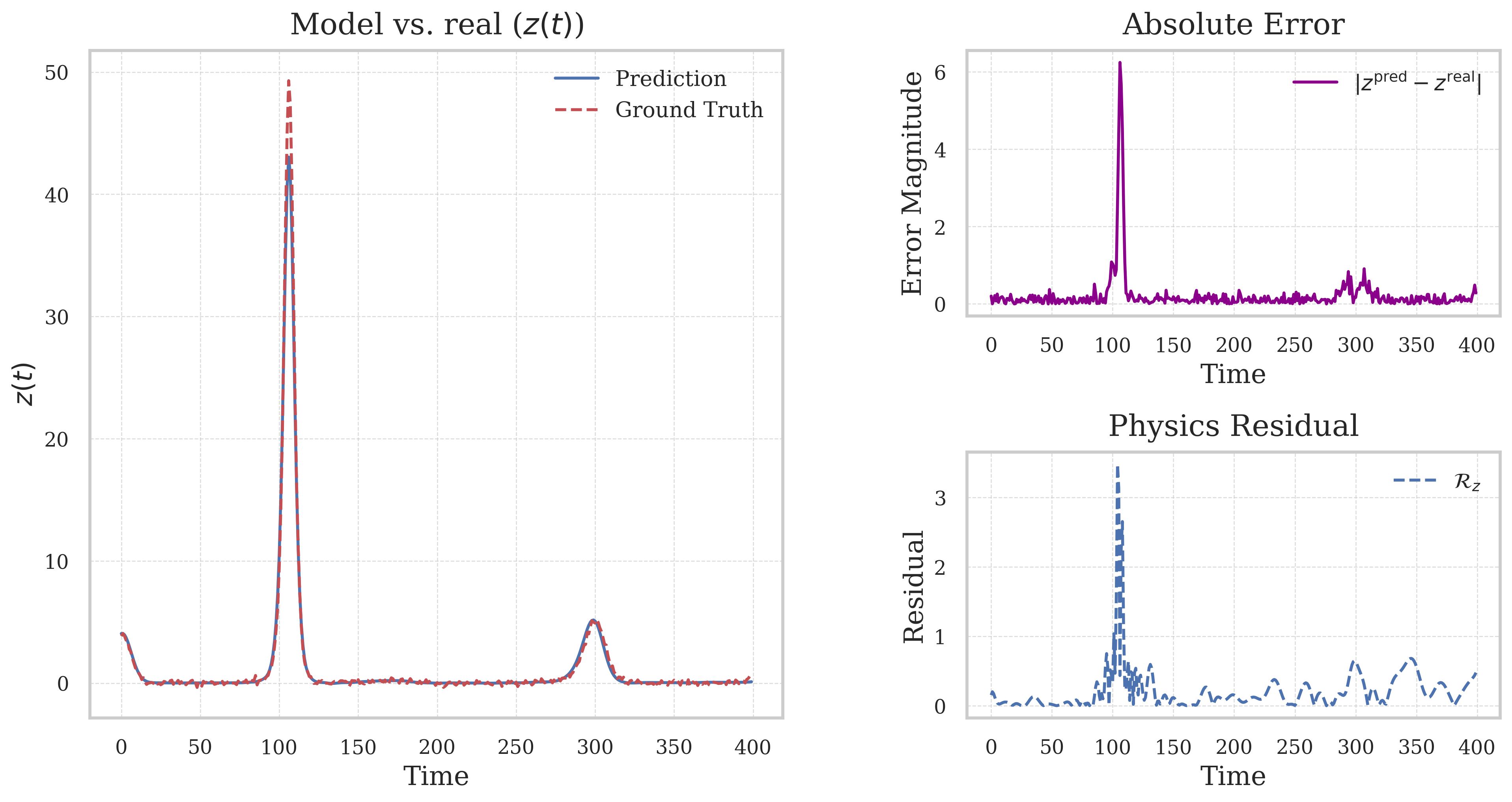}
    \caption{Comparison between the predicted and real trajectories of $z(t)$ for a test dataset (id = 19) sequence (left). The top right shows the corresponding absolute error, while the bottom right illustrates the physics residual $\mathcal{L}_{\text{ODE}_z}$ computed using PINN-inferred and identification bloc.}
    \label{Example_ross_physics_detection}
\end{figure}

We found that the model's prediction error is highly correlated with the ODE loss, with $R^2 > 0.7$, for both noisy and non-noisy data. This information could be used to provide a confidence level for the model's prediction; nevertheless, it does not indicate the amplitude of the error. This can be seen in the fact that the correlation is high for low loss $\mathbf{MAE_z}$ values (Fig.~\ref{r2ode_mae}), and decreases for high losses. In Figs.~\ref{Example_ross_physics_detection}  we give an example of the cross validation utility.\\
Based on these results, we found that the model is able to use noisy data with a few data points of variable lengths to capture the dynamics. An additional block could be used to cross validate the model's prediction. In the next section, we explore a 3D PDE example to study the model's ability to capture the physics of a multi-dimensional problem with parametric boundary conditions (BC).

\subsection{Application 2 : 2D Navier Stokes, flow around cylinder with parametric inlet function}\label{NS-App}
In this application, we tested the model 2D unsteady incompressible  laminar flow around a cylinder with a parametric inlet boundary condition. The flow is governed by the Navier-Stokes equations (Eq.~\ref{NS-equation}). The data generation and preprocessing are explained in Sec.~\ref{NS-App-DG}, and we present the model architecture and training details in Sec.~\ref{NS-App-MA}. Finally, we analyze the results in Sec.~\ref{NS-App-RD}.

\subsubsection{Data generation and preparation}\label{NS-App-DG}

We use a Finite Element Method (FEM) pre-defined code implemented in FEniCSx~\cite{dokken_ns} (based on the benchmark problem in~\cite{featflow}) to solve the incompressible Navier-Stokes equations for a laminar flow past a cylindrical obstacle. The governing equation is the incompressible Navier-Stokes equation (Eq.~\ref{NS-equation}), with parameters $\rho = 1$ (density) and $\mu = 0.001$ (dynamic viscosity). The boundary conditions are defined as follows in Eq.~\ref{eq-BCNS}:
\begin{equation}\label{eq-BCNS}
\begin{aligned}
\mathbf{u} &= \mathbf{u}_{\text{inlet}}(t, \mathbf{x}) \quad \text{on the inlet}, \\
\mathbf{u} &= [0, 0] \quad \text{on the walls and obstacle (no-slip condition)}, \\
p &= 0 \quad \text{on the outlet}.
\end{aligned}
\end{equation}

The inlet velocity have a parabolic profile, with two parameters ($U_{\text{max}}$, $f$), as given in Eq.~\ref{eq-INNS}:
\begin{equation}\label{eq-INNS}
\begin{aligned}
\mathbf{u}_{\text{inlet}}(t, \mathbf{x}) &=  \left [4 U(t) \frac{y (H - y)}{H^2}, 0 \right] ,\\
U(t) &= U_{\max} t \left( \sin \left( \frac{\pi t}{f} \right)^2 + 1 \right).
\end{aligned}
\end{equation}
, and the initial conditions are zero:
\begin{equation}\label{eq-ICNS}
\mathbf{u}(0, \mathbf{x}) = 0, \quad p(0, \mathbf{x}) = 0, \quad \forall \mathbf{x} \in \Omega.
\end{equation}

The domain is a rectangle of length $L = 0.55$ and height $H = 0.205$, with a cylindrical obstacle of radius $r = 0.025$ located at $(c_x, c_y) = (0.1, 0.1)$. We used a quadrilateral structured mesh to ensure good refinement near the obstacle, with a total of 9491 nodes, ensuring accurate resolution of the flow features (Fig.~\ref{NS-geometry}). The simulation's total duration is $T = 6$ with a sampling rate of $\Delta T = 0.03$.
The used code~\cite{dokken_ns} employs the first-order backward difference scheme with a Crank-Nicolson discretization in time, along with a semi-implicit Adams-Bashforth approximation for numerical stability. We also tested different mesh levels to ensure the convergence of the simulation.

\begin{figure}[h]
\centering
\includegraphics[width=0.7\textwidth]{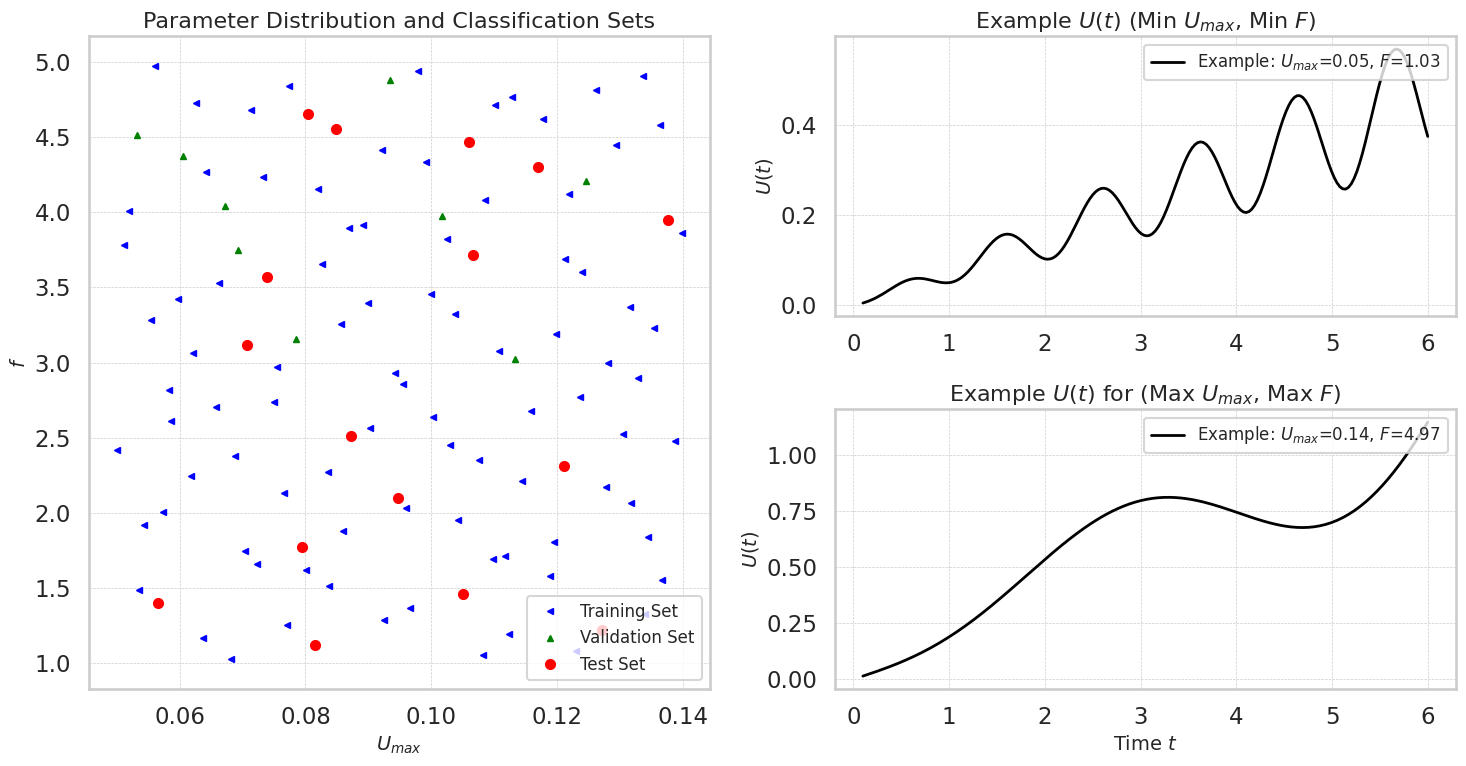}
\caption{The distribution of inlet velocity parameters $U_{\max}$ and $f$ used for training, validation, and testing on the left, along with two extreme examples on the right of the $U(t)$ function.}
\label{fig:distribution}
\end{figure}

Multiple simulations are conducted with different values of $U_{\text{max}}$ and $f$ sampled from a predefined distribution. The distribution of these parameters is shown in Fig.~\ref{fig:distribution}. We used 90 datasets for training, validated the model on a small range of $f > 3$ using 9 datasets, and tested the model on 16 datasets. A small value of $f$ results in highly oscillatory behavior, while a larger value give less oscillation (Fig.~\ref{fig:distribution}, right).

\subsubsection{Model architecture and training process}\label{NS-App-MA}

\begin{figure}[h!]
\centering
\includegraphics[width=0.8\textwidth]{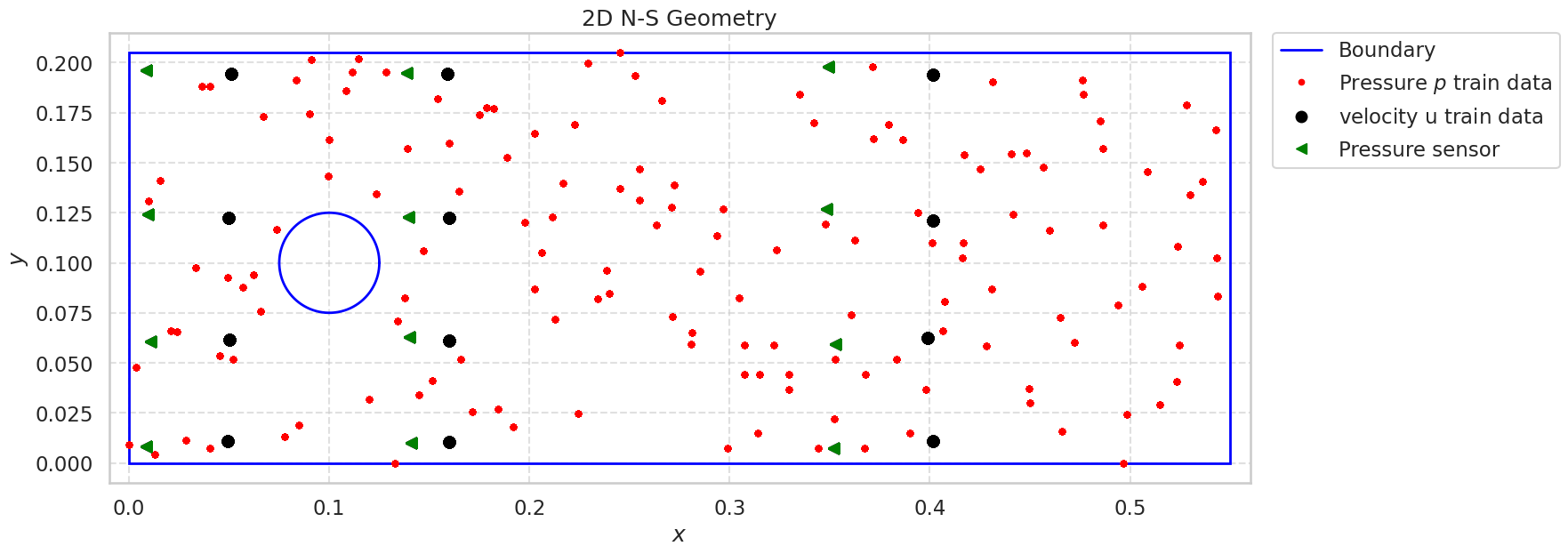}
\caption{The distribution of the 12 pressure sensors, $p_i$, where $i \in [1,12]$, used as input for the Deep Set model $X_{\text{DS}}$, is shown, along with the distribution of the velocity and pressure points used for the supervised loss $\mathcal{L}_{\text{SUP}}$ given in Eq.~\ref{loss_NS}.}
\label{NS-geometry}
\end{figure}

The architecture used in this application is similar to the one used for the Rössler ODE (see Sec.~\ref{Model architecture and training}). In this case, the architecture includes an identification block to predict the parameters of the inlet function, $U_{\max}$ and $f$. The model takes as input $\textbf{X}_{\text{DS}}$ (Eq.~\ref{eq-inpds}).
\begin{equation}\label{eq-inpds}
\textbf{X}_{DS} = \{t,~p_1,~\cdots,~p_i,~\cdots,~p_{12}\}\in\mathbb{R}^{b\times N_{DS} \times 13} \quad i\in[2,11] \quad \text{and} \quad t \in[0,~ 0.6\times T]\\
\end{equation}
The input data $\textbf{X}_{DS}$ represents time and pressure data from 12 sensors distributed at 3 locations, as described in Fig.~\ref{NS-geometry}. In this application, we used a fixed window length, hence $N_{DS}=120$ snapshots (the simulation time is $T=6.0$ with a sampling rate of $\Delta T=0.03$). The Deep Set uses this data to calculate a feature vector $F_{DS}$, which is then used by the identification block to predict the inlet's function parameters $U(t)$, specifically $U_{\max}$ and $f$. This same vector is used as input for a post-encoding MLP, which calculates another feature vector $F_{\text{PINN}}$, which is then concatenated with time and space data to predict the velocity vector $\mathbf{u} = [u_x, u_y]$ and pressure $p$.

To train the model, we used the following loss function, as given in Eq.~\ref{loss_NS}.
\begin{equation}\label{loss_NS}
\begin{aligned}
\mathcal{L}_{\text{total}} &= \alpha\mathcal{L}_{\text{SUP}}  + \lambda \mathcal{L}_{\text{PDE}} + \eta (\mathcal{L}_{\text{ICs}} +  \mathcal{L}_{\text{BCs}}) + \mathcal{L}_{\text{P}}, \\
\mathcal{L}_{\text{Sup}} &= \|\textbf{u}(t, \textbf{x}) - \widetilde{\textbf{u}}(t, \textbf{x})\|_2 + \|p(t, \textbf{x}) - \widetilde{p}(t, \textbf{x})\|_2 + \|\textbf{u}(t, \textbf{x}_{inlet}) - \widetilde{\textbf{u}}(t, \textbf{x}_{inlet})\|_2, \\
\mathcal{L}_{\text{ICs}} &= \|\textbf{u}(0, \textbf{x}) - \widetilde{\textbf{u}}(0, \textbf{x})\|_2 + \|p(0, \textbf{x}) - \widetilde{p}(0, \textbf{x})\|_2,\\
\mathcal{L}_{\text{BCs}} &= \| \textbf{u}(t, \textbf{x}_{wall}) - \widetilde{\textbf{u}}(t, \textbf{x}_{wall}) \|_2 + \| \textbf{u}(t, \textbf{x}_{cyl}) - \widetilde{\textbf{u}}(t, \textbf{x}_{cyl}) \|_2  \\
\mathcal{L}_{\text{PDE}} &= \|\mathcal{L}_{eq1}^x\|_2 + \|\mathcal{L}_{eq1}^y\|_2 + \|\mathcal{L}_{eq2}\|_2 \\
\mathcal{L}_{\text{P}} &= \|U_{max} -  \widetilde{U}_{max}\|_2 + \|f -  \widetilde{f}\|_2
\end{aligned}\\
\end{equation}
, where $\widetilde{.}$ denotes the known data. The first part, $\mathcal{L}_{\text{Sup}}$, is the supervised loss of the training where $\lambda = 10$, which also includes the inlet loss. We used a limited number of velocity points (12 points) and 147 pressure points chosen randomly, using the total time snapshots, hence $t \in [0, T]$. The distribution of pressure and velocity points used for the supervised loss is given in Fig.~\ref{NS-geometry}. For the inlet function, we used the real parameters $(\widetilde{U}_{\text{max}}, \widetilde{f})$ to calculate the real inlet velocity $\widetilde{\textbf{u}}_{\text{inlet}}$.

The second part, $\mathcal{L}_{\text{ICs}}$, represents the initial conditions. In this application, at $t=0$, velocity and pressure are set to 0, with $\eta = 0.1$. We used $\mathcal{L}_{\text{BCs}}$ to impose the non-slip boundary condition for the wall and cylinder (annotated as $cyl$ in the equation). The physics loss $\mathcal{L}_{\text{PDE}}$ combines the residuals of the projections in the $x$ and $y$ directions for the conservation of momentum, $\mathcal{L}_{\text{eq1}}^x$ and $\mathcal{L}_{\text{eq1}}^y$, respectively, and the conservation of mass equation, $\mathcal{L}_{\text{eq2}}$ (Eqs.~\ref{NS-equation}), which is used to calculate the velocity and pressure at positions where we do not have any supervised data. At the beginning of the training, $\lambda = 0$, and we then increase it depending on the $\mathcal{L}_{\text{Sup}}$. The last part of the loss function, $\mathcal{L}_{\text{P}}$, represents the parameter loss of the inlet velocity function. For training, we used a batch size of 15 datasets. The validation dataset was used to study the convergence and select an optimal model. We used a similar training approach as Algo.~\ref{alog_app1}. At the start of training, the learning rate is set to $lr = 10^{-3}$, with $\alpha = 10$, $\lambda = 0$, and $\eta = 0.1$. The learning rate $lr$ is decreased over epochs, with multiple thresholds adjusted to optimize training speed. After a pre-training phase where $\text{epoch} > \text{threshold}_2$, the physics loss is activated, and trainable variables are restricted to only the PINN model. The physics loss weight $\lambda$ is gradually increased when the supervised loss drops below a defined threshold.

The model architecture is given in Tab.~\ref{tab:architecture_app2}.

\begin{table}[h]
\centering
\renewcommand{\arraystretch}{1.3}
\setlength{\tabcolsep}{6pt} 
\caption{Architecture details of the proposed model for the Navier-Stokes model.}\label{tab:architecture_app2}
\begin{tabular}{|l|c|c|c|c|p{5cm}|}
\hline
\textbf{Model} & \textbf{Activation Function} & \textbf{Layers} & \textbf{Width} & \textbf{Output Dim} & \textbf{Comments} \\ 
\hline
\text{Deep set} & $\sin(3.14 X) \cdot \text{swish}(X)$ & 4 & 121 & 121 & Aggregation is performed using summation. \\ 
\hline
\text{Identification} & $\sin(3.14 X) \cdot \text{swish}(X)$ & 4 & 121 & 2 & Outputs the identified parameters. \\ 
\hline
\text{Post Deep Set} & $\sin(3.14 X) \cdot \text{swish}(X)$ & 2 & 121 & 5 & Takes the Deep Set feature vector as input and produces another feature vector that is fed to the PINN. \\ 
\hline
\text{PINN} & $\sin(3.14 X) \cdot \text{swish}(X)$ & $5\times2$ & $128\times2$ & 3 & The model takes as input the Post Deep Set output along with spatial and temporal vectors. We have two PINN models, and the outputs of these two PINN models are summed to give the prediction. We found this approach to be more accurate.  \\ 
\hline
\end{tabular}
\end{table}

\begin{figure}[H]
\centering
\includegraphics[width=1.\textwidth]{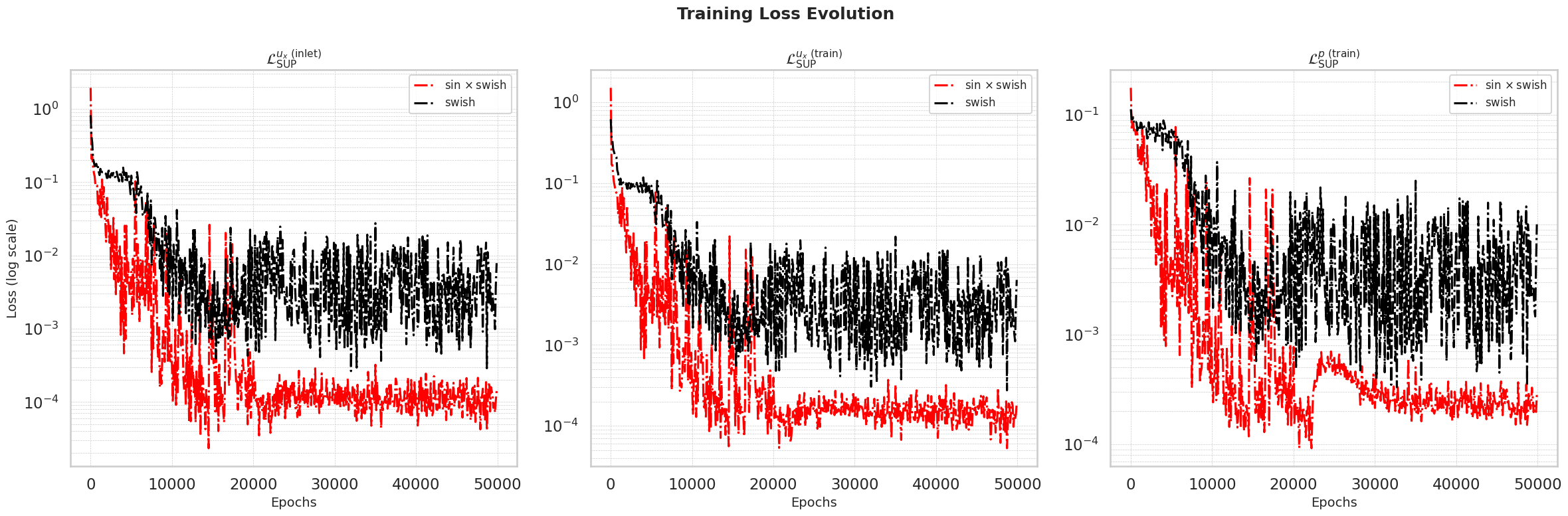}
\caption{The differences in the loss convergence using the same architecture but with different activation functions. On the left, we have the inlet loss function, in the center the supervised loss of $u_x$, and on the right, the supervised loss of $p$. }
\label{activation_func}
\end{figure}

As seen in Sec.~\ref{Model architecture and training}, we used a special activation function that combines a sinusoidal function with the Swish function. The motivation is simply to accelerate the training process (Fig.~\ref{activation_func}). We found that this function fits easily with outputs that exhibit sinusoidal behavior (we did the same in Sec.~\ref{ODE-App}).

\subsubsection{Results and discussion}\label{NS-App-RD}
To assist the model performance we used the test datasets (unseen data during the training), we selected an optimal model based on it performance on supervised loss and also on the physics residual of the validation datasets. We used the \textbf{MAE} metric where we compared the model prediction with the FEM simulation (high fidelity data), on the whole domain $x,y \in \Omega $ at $N$ snapshots, where $t > 0.6T$, which mean we tested the model on time snapshot that is not feed to the Deep set via $X_{DS}$ (Eq.~\ref{eq-inpds}). We first calculated for each test dataset the \textbf{MAE}, median-\textbf{Absolute Error (AE)}, and 90\%-\textbf{AE} over the whole spatial domain for every snapshot, given in Eq.~\eqref{eq-error-metrics}:

\begin{equation}\label{eq-error-metrics}
\begin{aligned}
    \mathcal{E}_{\text{MAE}}(t) &= \frac{1}{N} \sum_{i=1}^{N_{x,y}} \left| f_i^{\text{pred}}(t) - f_i^{\text{true}}(t) \right|, \\
    \mathcal{E}_{\text{median-AE}}(t) &= \operatorname{median} \left( \left| f_i^{\text{pred}}(t) - f_i^{\text{true}}(t) \right| \right), \quad i = 1, \dots, N_{x,y}, \\
    \mathcal{E}_{90\%-\text{MAE}}(t) &= P_{90} \left( \left| f_i^{\text{pred}}(t) - f_i^{\text{true}}(t) \right| \right), \quad i = 1, \dots, N_{x,y}.
\end{aligned}
\end{equation}
Where \( f_i^{\text{pred}}(t) \) and \( f_i^{\text{true}}(t) \) are the predicted and true values of velocity components \( u_x, u_y \) or pressure \( p \) at spatial point \( i \), \( N_{x,y} \) is the total number of spatial points, and \( P_{90} \) represents the 90th percentile of the absolute error distribution. Next, we computed the time mean values of these metrics for each test dataset (Eq.~\ref{eq-time-avg}):

\begin{equation}\label{eq-time-avg}
\bar{\mathcal{E}} = \frac{1}{N_t} \sum_{t=1}^{N_t} \mathcal{E}(t_i),
\end{equation}
Where \( N_t \) is the total number of time snapshots. The results are presented in Tab.~\ref{tab:results_NS_all}, where each row corresponds to an aggregated value (mean, median, and 90th percentile) computed across all test datasets for all metrics. 

\begin{table}[h]
\centering
\caption{Aggregated model prediction error for \( u_x \), \( u_y \) and \( p \), evaluated using : \( \bar{\mathcal{E}}_{\text{MAE}} \),  \( \bar{\mathcal{E}}_{\text{median-AE}} \), and \( \bar{\mathcal{E}}_{\text{90\%-AE}} \). The table represent the distribution of these metrics across all test datasets, summarized by their mean, median (50\%), and 90th percentile.}
\label{tab:results_NS_all}
\begin{tabular}{llccc}
\toprule
\textbf{Variable} & \textbf{Aggregation} & $\bar{\mathcal{E}}_{\text{MAE}}$ & $\bar{\mathcal{E}}_{\text{median-AE}}$ & $\bar{\mathcal{E}}_{\text{90\%-AE}}$ \\
\midrule
\multirow{3}{*}{\( u_x \)} 
  & Mean   & 0.028 & 0.015 & 0.068 \\
  & 50\%   & 0.026 & 0.012 & 0.067 \\
  & 90\%   & 0.035 & 0.018 & 0.084 \\
\midrule
\multirow{3}{*}{\( u_y \)} 
  & Mean   & 0.013 & 0.005 & 0.034 \\
  & 50\%   & 0.013 & 0.005 & 0.034 \\
  & 90\%   & 0.016 & 0.007 & 0.038 \\
\midrule
\multirow{3}{*}{\( p \)} 
  & Mean   & 0.013 & 0.010 & 0.025 \\
  & 50\%   & 0.007 & 0.003 & 0.016 \\
  & 90\%   & 0.021 & 0.016 & 0.040 \\
\bottomrule
\end{tabular}
\end{table}

The model achieves high performance as indicated by the \textbf{MAE} metric. From these results, we can see that the model successfully identified the boundary condition’s parameters and use them to predict velocity, even when they are not explicitly provided as input. While the error could be minimized by adding some supervised data and including physics residual points where the error is high, as given in~\citep{HANNA2022115100}.

\begin{figure}[H]
\centering
\includegraphics[width=0.8\textwidth]{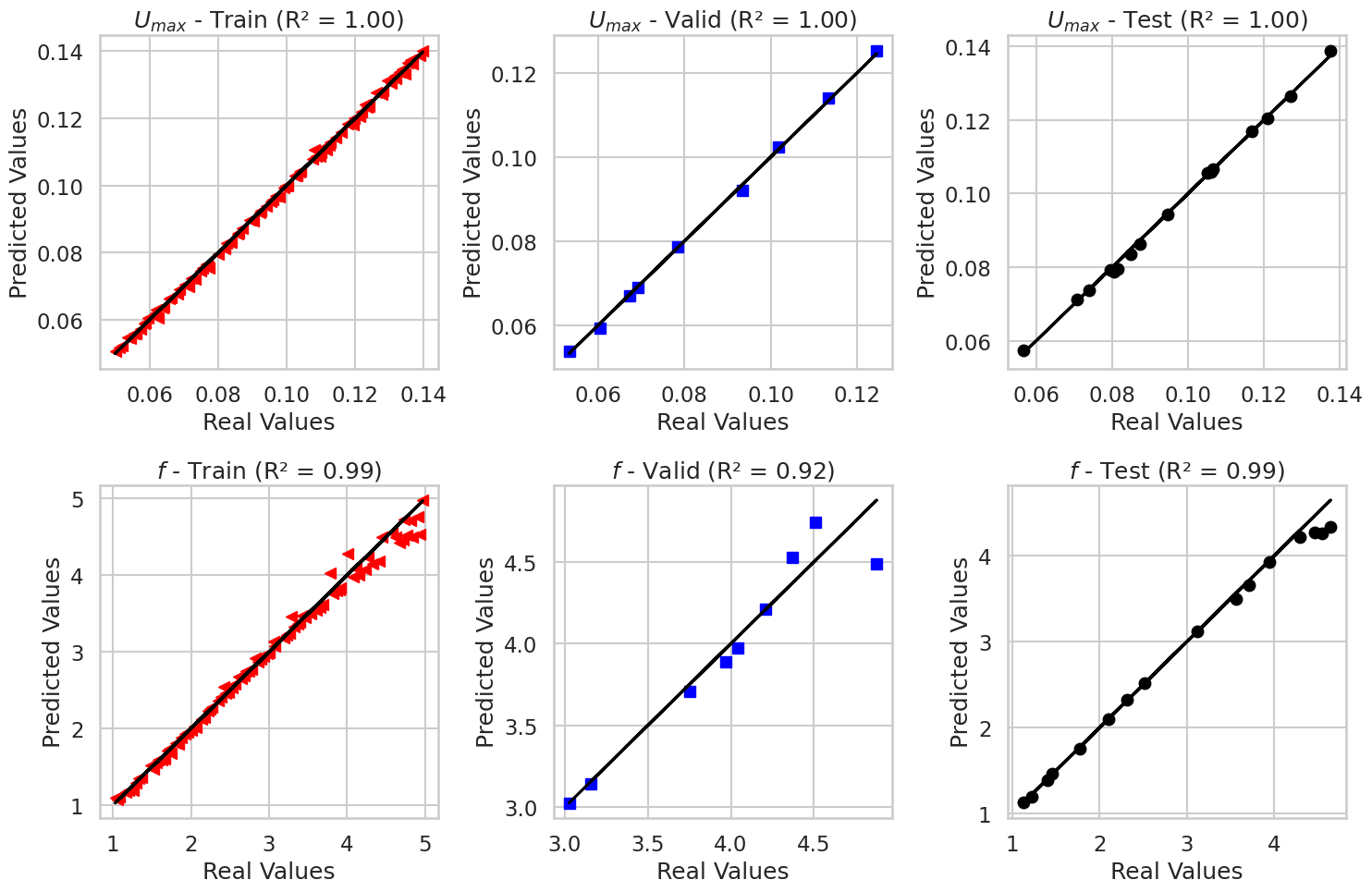}
\caption{The accuracy the identification blocks to predict the inlet function parameters $U_{max}$ (at the Top) and $f$ (at the Bottom), for train, validation and test datasets.}
\label{umax_f_results}
\end{figure}

We also tested the model's prediction of the parameters of the inlet function \( u_{\text{inlet}} \). In Fig.~\ref{umax_f_results}, we can see that the accuracy of the model using (\( R^2 \)) metric is good ($R^2  > 0.9$). We can also use this block to validate the prediction of the PINN by comparing the inlet function from the PINN and the prediction from the identification block.

To test the possibility of using cross validation, we tested the model's prediction of \( u_x^{\text{inlet}} \) (Eq.~\ref{eq-INNS}). We calculated the scaled-\textbf{MAE} of the model prediction (Eq.~\ref{Scaled-MAE-model}) of \( u_x^{\text{inlet}} \) and the constructed \( \widetilde{u}_x^{\text{inlet}} \) from the identification block, and the scaled-\textbf{MAE} of \( \widetilde{u}_x^{\text{inlet}} \) and the constructed \( \widehat{u}_x^{\text{inlet}} \) from the real parameters (Eq.~\ref{Scaled-MAE-comparison}).

\begin{equation}\label{Scaled-MAE-model}
\text{Scaled-MAE}_{\text{model}} = \frac{1}{\max(\widetilde{u}_x^{\text{inlet}}) - \min(\widetilde{u}_x^{\text{inlet}})} \cdot \sum \left| u_x^{\text{inlet}} - \widetilde{u}_x^{\text{inlet}} \right|,
\end{equation}

\begin{equation}\label{Scaled-MAE-comparison}
\text{Scaled-MAE}_{\text{comparison}} = \frac{1}{\max(\widetilde{u}_x^{\text{inlet}}) - \min(\widetilde{u}_x^{\text{inlet}})} \cdot \sum \left| \widehat{u}_x^{\text{inlet}} - \widetilde{u}_x^{\text{inlet}}  \right|
\end{equation}

From these results, we can see that cross validation is efficient in detecting bad predictions of the model with \( R^2 = 0.79 \). We note here that we selected an optimal model based only on the accuracy of the prediction of the PINN and identification block over the validation dataset, without any criteria for cross validation of these two blocks. We believe that these results can be improved by choosing a compromise between the model's accuracy and the efficiency of detecting bad predictions.

\begin{figure}[H]
\centering
\includegraphics[width=.7\textwidth]{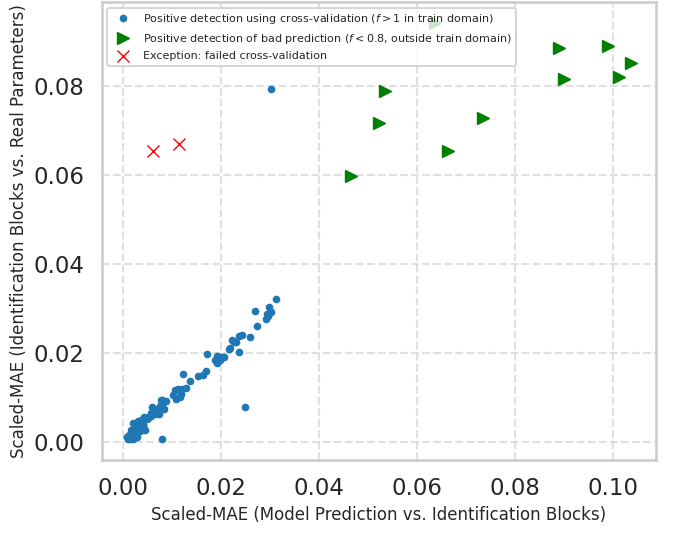}
\caption{Correlation between the scaled-MAE errors of the model. The x-axis represents the scaled-MAE between the model prediction of $u_x^{\text{inlet}}$ and the reconstructed $\widetilde{u}_x^{\text{inlet}}$ from the identification blocks, while the y-axis represents the scaled-MAE between $\widetilde{u}_x^{\text{inlet}}$ and the reconstructed $\widehat{u}_x^{\text{inlet}}$ from the real parameters. The coefficient of determination is $R^2 = 0,79$. The red crosses represent exceptional cases where the cross-validation fails, while the other markers indicate successful detections within and outside the training domain.}
\label{cross_valid_ident_NS}
\end{figure}

\subsection{Application 3 : 1D heat problem, application to real data}\label{HT-App}
In this application, we tested the model on real data from the heating of composite materials made of a thermoplastic polymer and glass fibers using an InfraRed (IR) oven. In this case, the model is trained only on noisy simulation data and then tested on real data.

\subsubsection{Samples}
\begin{figure}[H]
\centering
\includegraphics[width=0.7\textwidth]{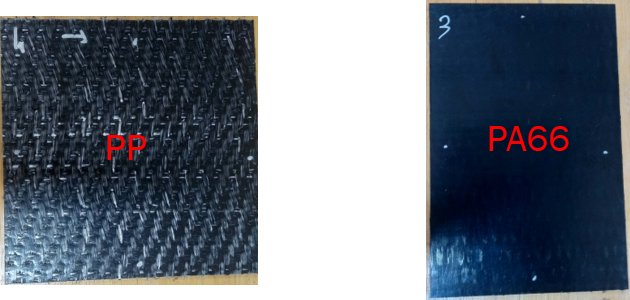}
\caption{Pictures showing the surface of the PP glass fibers plate on the left, where we can see that the quality of the surface is variable (which implies a variation of absorbed heat flux during the heat process). On the right is the PA66-glass fibers plate, where the surface quality is more homogeneous. We took measurements of thickness at 5 different locations, as demonstrated in the picture (4 white points in edges) and in the center.}
\label{composite_PA_PP}
\end{figure}

We used two types of composite plates in this study: a blend of Polyamide (PA66) and glass fibers, and a blend of  Polypropylene (PP) also reinforced with glass fibers. The Fig.~\ref{composite_PA_PP} shows PA66-glass fiber plates on the right and PP-glass fiber on the left. The material properties of each constituent of the composites are given in Tab.~\ref{tab:material_properties}.\\

\begin{table}[h]
\caption{Material properties of E-glass Fiber, PP (Isotactic Polypropylene), and PA66.}
    \label{tab:material_properties}
    \centering
    \renewcommand{\arraystretch}{1.2} 
    
    \begin{tabular}{l c c c c}
        \hline
        \textbf{Property} & \textbf{E-glass} & \textbf{PP} & \textbf{PA66} & \textbf{Units} \\
        \hline
        Density ($\rho$) & 2580 & 910 & 1140 & kg/m$^3$ \\
        Specific Heat Capacity ($C_p$) & 830 & 1610 & 1460 & J/(kg$\cdot$K) \\
        Thermal Conductivity ($\lambda$) & 1.0 & 0.22 & 0.2 & W/(m$\cdot$K) \\
        \hline
    \end{tabular}
    
\end{table}

We used 10 PP plates and 2 PA plates. The nominal geometries is 15.3~cm $\times$ 15.3~cm $\times$ 2.4~mm for the PA66 plates, and 15.3~cm $\times$ 10.1~cm $\times$ 1.5~mm for the PP plates, corresponding to length, width, and thickness, respectively.

\subsubsection{Equipment}
\begin{figure}[H]
\centering
\includegraphics[width=0.75\textwidth]{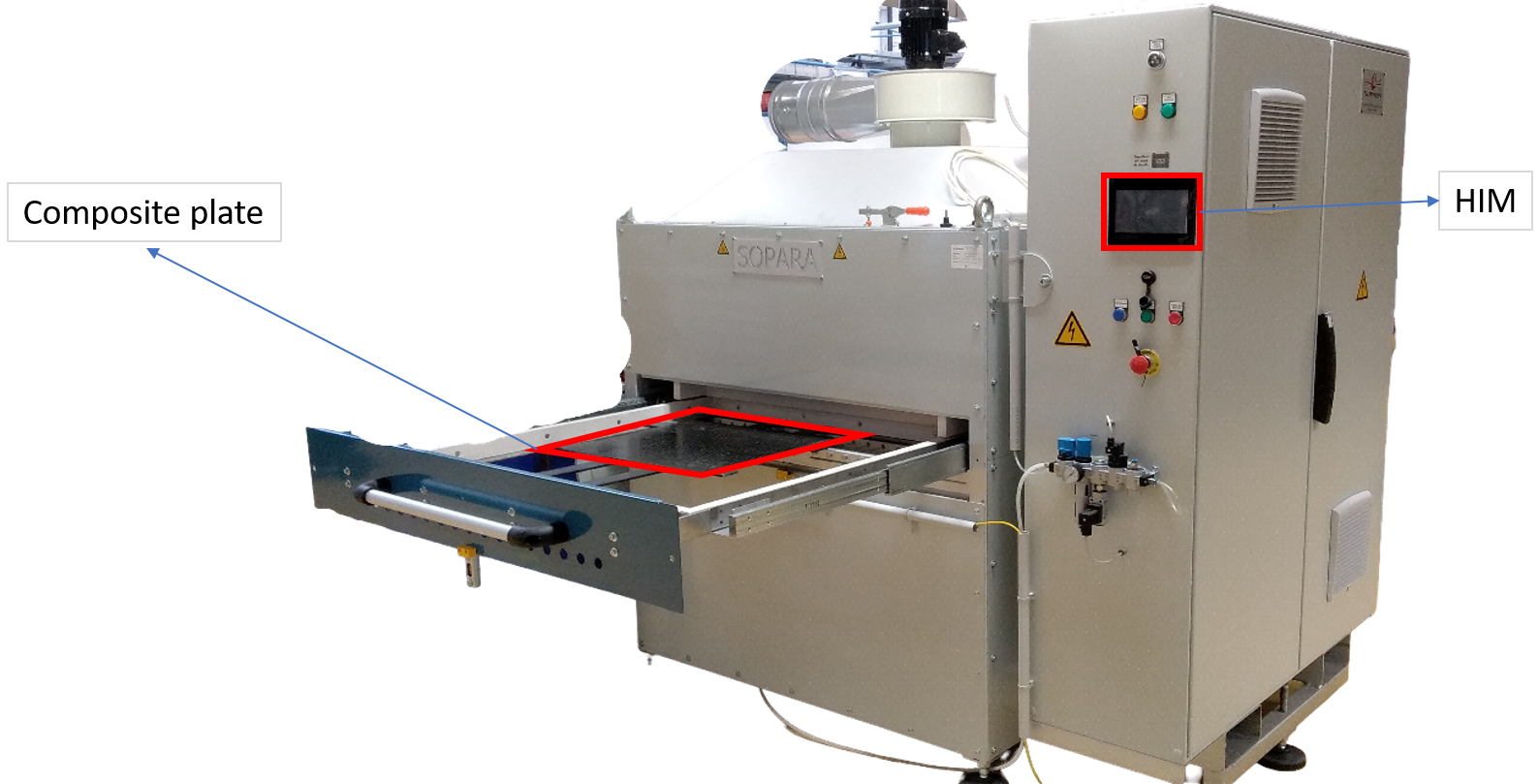}
\caption{Picture of the IR oven used to generate experimental data by heating PA and PP plates.}
\label{IR_oven}
\end{figure}

\begin{figure}[H]
\centering
\includegraphics[width=0.6\textwidth]{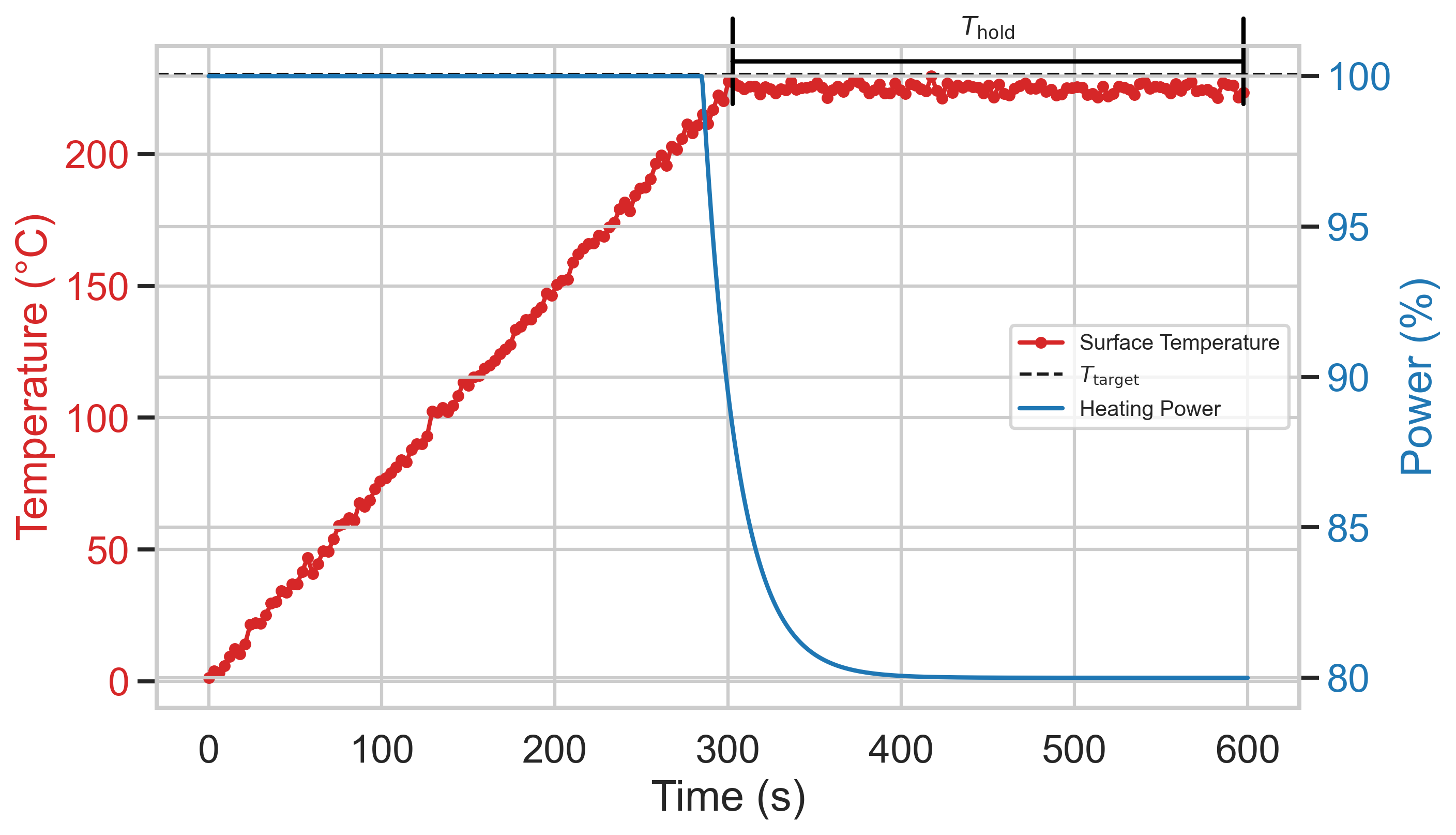}
\caption{This graph illustrates the qualitative behavior (does not reflect real data) of temperature ($T_{\text{Top}}$) and power ($P_{\text{Top}}$) in the infrared oven during heating. The temperature rises to $T_{\text{target}}$ at $t_{\text{target}}$, after which the system holds the temperature at $T_{\text{hold}}$.}
\label{oven_temperature_power_plot}
\end{figure}

The heating of these plates is performed using an infrared oven (SOPARA, Fig.~\ref{IR_oven}). Inside the oven, two symmetric pyrometers measure the surface temperature of the plates at the top and bottom during heating, along with an ambient air temperature sensor. The oven is equipped with an automation system that allows adjustment of its parameters: the percentage of heating power at the top ($P_{top}$) and bottom ($P_{bottom}$), the target temperature ($T_{target}$), and the hold time ($T_{hold}$), as described in Fig.~\ref{oven_temperature_power_plot}.\\

\begin{figure}[H]
\centering
\includegraphics[width=.7\textwidth]{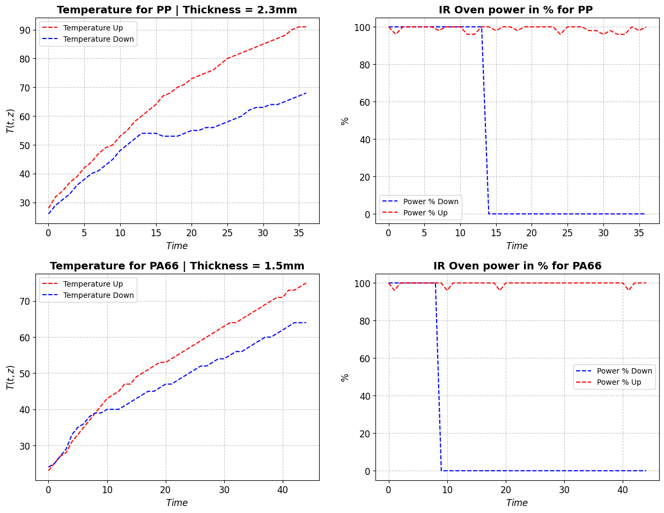}
\caption{Examples of data collected during the heat process of two plates (PP at the top, and PA at the bottom). On the right, we have the temperature at the top and bottom during the heating process, and on the left, we have the power of the heating at the top and bottom. As seen, we used a special protocol, where we set the power at the bottom to 0\% at a random instant \( T_{pdb} \).}
\label{Example_real_data}
\end{figure}

During the heating, the automate logs heating data w.r.t time (Fig.~\ref{Example_real_data}), namely temperature (in $°C$) at the top and bottom also the oven air temperature, the \% of heating power at the Top and at the Bottom. The acquisition frequency is set to 1 Hz, nevertheless in practice this frequency varies between 0.2 to 1 Hz. We also note that the quality of the pyrometer  sensors is susceptible to some variation (for example, because of the color or the surface quality of the plates). 

\subsubsection{Testing scenario}
The heating scenario used to heat this plates is the following (Examples in Fig.~\ref{Example_real_data}):
\begin{itemize}
\item The oven is set to heat both sides (Top and Bottom) using 100\% of it power,
\item At a random time $T_{pdb}$, the heat power at the bottom is set to 0\%, while the heating continue in the Top with 100\% of power.
\end{itemize}

Where $T_{pdb}$ varies from 4 s to 16 s, we used the same heating scenario for both PA66 plates and PA plates. This setup was chosen to challenge the model's ability to adapt to sudden changes in boundary conditions.

\subsubsection{Numerical data generation}
\begin{figure}[H]
\centering
\includegraphics[width=0.7\textwidth]{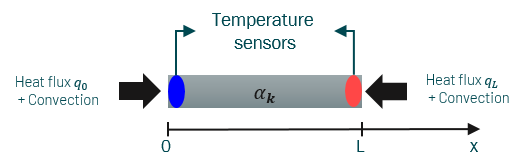}
\caption{Simplified 1D schematic of the simulation of the heat process, where $\alpha_k$ represents the diffusivity of the $k$ test case, and the red and blue circles represent temperature sensors, while the arrows represent the BC flux $q_0$ and $q_L$ and convection.}
\label{graph_heating_1d}
\end{figure}

To train the model, we generated synthetic data based on the variations observed in the experimental data. Since the thickness of the plates is small compared to the length and width of the plates, the heating process can be simplified to a 1D heat problem, where heat transfer in the thickness direction is the most significant. The simplified 1D heat problem is illustrated in Fig.~\ref{graph_heating_1d}.\\

Based on these data, we generated $N = 2640$ datasets of 1D heat conduction (Eq.~\ref{HT-eq}), with variables heat flux (BCs), thickness, thermal properties (in this study, the diffusivity variation is not significant), and the initial condition. We also introduced a variation in the heating process, which is the drop in the percentage of power applied to the bottom at a random time $T_{pdb}$. These variations are described in Tab.~\ref{tab:variable_ranges}. We used Sobol  to generate a combination of all these variations, considering that a dataset size of $N = 2640$ is not very large.

\begin{table}[h]
    \centering
    \caption{The ranges of variables for heat flux, thickness, thermal properties, initial conditions, and the time $T_{pdb}$ for the drop in the percentage of power are provided with descriptions.}
    \label{tab:variable_ranges}
    \renewcommand{\arraystretch}{1.3}
    \begin{tabular}{l p{4cm} l}
        \toprule
        Variable & Range & Comment \\
        \midrule
        Diffusivity ($10^{-7}~\mathrm{m}^2/\mathrm{s}$) & 1.48 -- 1.52 & Varies slightly in a narrow range. \\
        $q/k$ ($10^3$~K$\cdot$m$^{-1}$) & $9.6-10.6$ (low) & Two distinct variation ranges. \\
        &$16.3-18.3$ (high) & \\
        $L$ (mm) & 1.58 -- 2.41 & Linearly increasing. \\
        IC (°$C$) & 22 -- 30 & Randomly generated within the range.\\
        $T_{\text{pdb}}$ (s) & 4 -- 15 & Linearly increasing. \\
        \bottomrule
    \end{tabular}
\end{table}

\begin{figure}[H]
\centering
\includegraphics[width=0.8\textwidth]{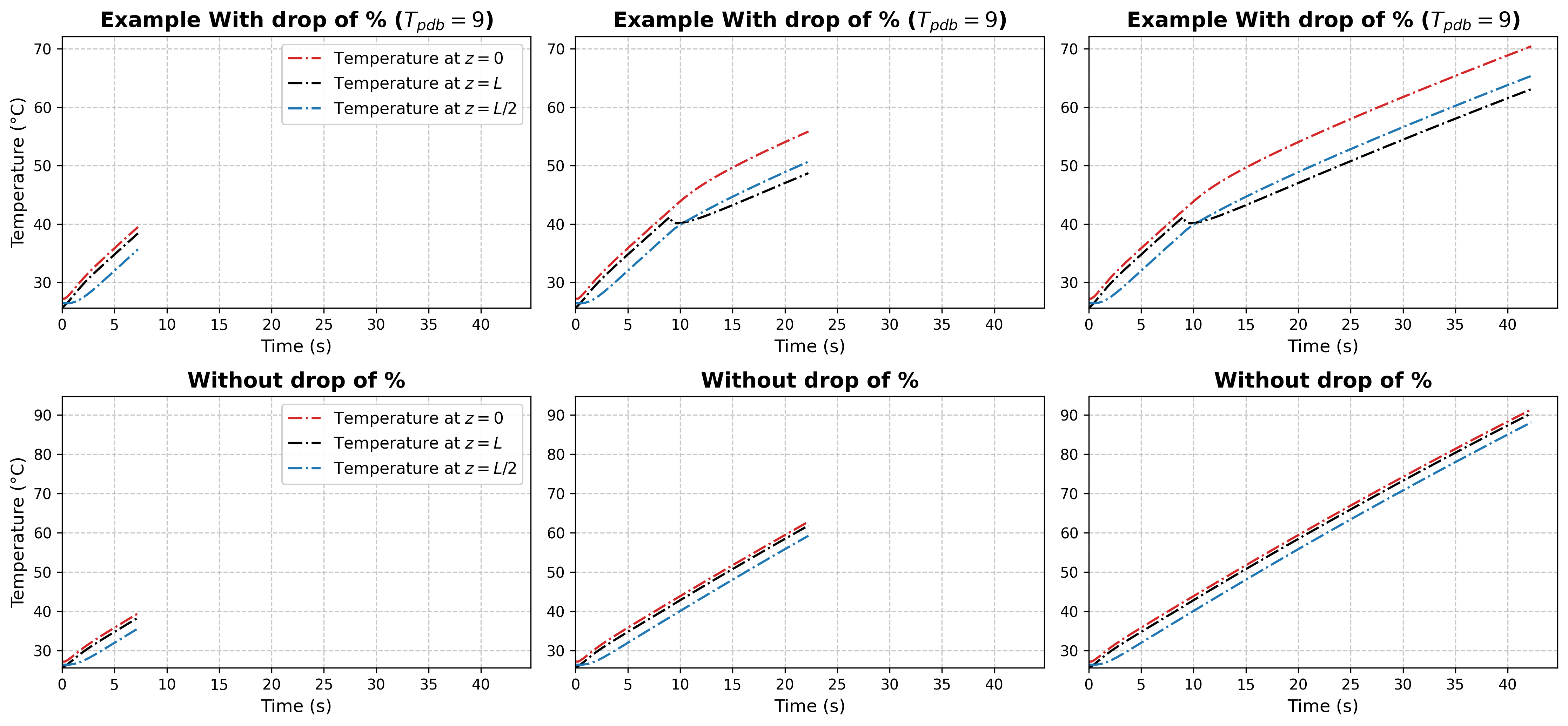}
\caption{An example of two datasets with the same simulation parameters (ICs/BCs, thickness, etc.) is shown: in the top figures, the first dataset include a drop in power at the bottom boundary to 0\%; in the bottom figures, the second dataset do not include this power drop.}
\label{simu_data_example}
\end{figure}

For each dataset (combination of the variations in Tab.~\ref{tab:variable_ranges}), we simulate the 1D heating problem (Eq.~\ref{HT-eq}) using the backward Euler scheme, where $h = 8$ is the natural thermal convection coefficient. We study the convergence of the numerical solution by decreasing both the time step $\Delta t$ and the spatial step $\Delta x$. For each dataset, we generate both cases: one where there is a drop in the power heating percentage at the bottom ($x = L$), and one without the drop in power (Fig.~\ref{simu_data_example}). This means that the model is trained in both cases. We did this to give the model the ability to learn the impact of the drop in heat power percentage. For each simulation, we use $t \in [0,~T]$, where $T = 45~\text{s}$, while for the ICs, as given in Tab.~\ref{tab:variable_ranges}, for each dataset, we generate $T(t = 0,~x = 0)$ and $T(t = 0,~x = L)$, and then we calculate the ICs $T(t = 0,~x)$ for each $x \in ]0, L[$ using an affine function based on $T(x=0, t)$ and $T(x=L, t)$.

\subsubsection{Model architecture and training process}
\begin{figure}[H]
\centering
\includegraphics[width=0.75\textwidth]{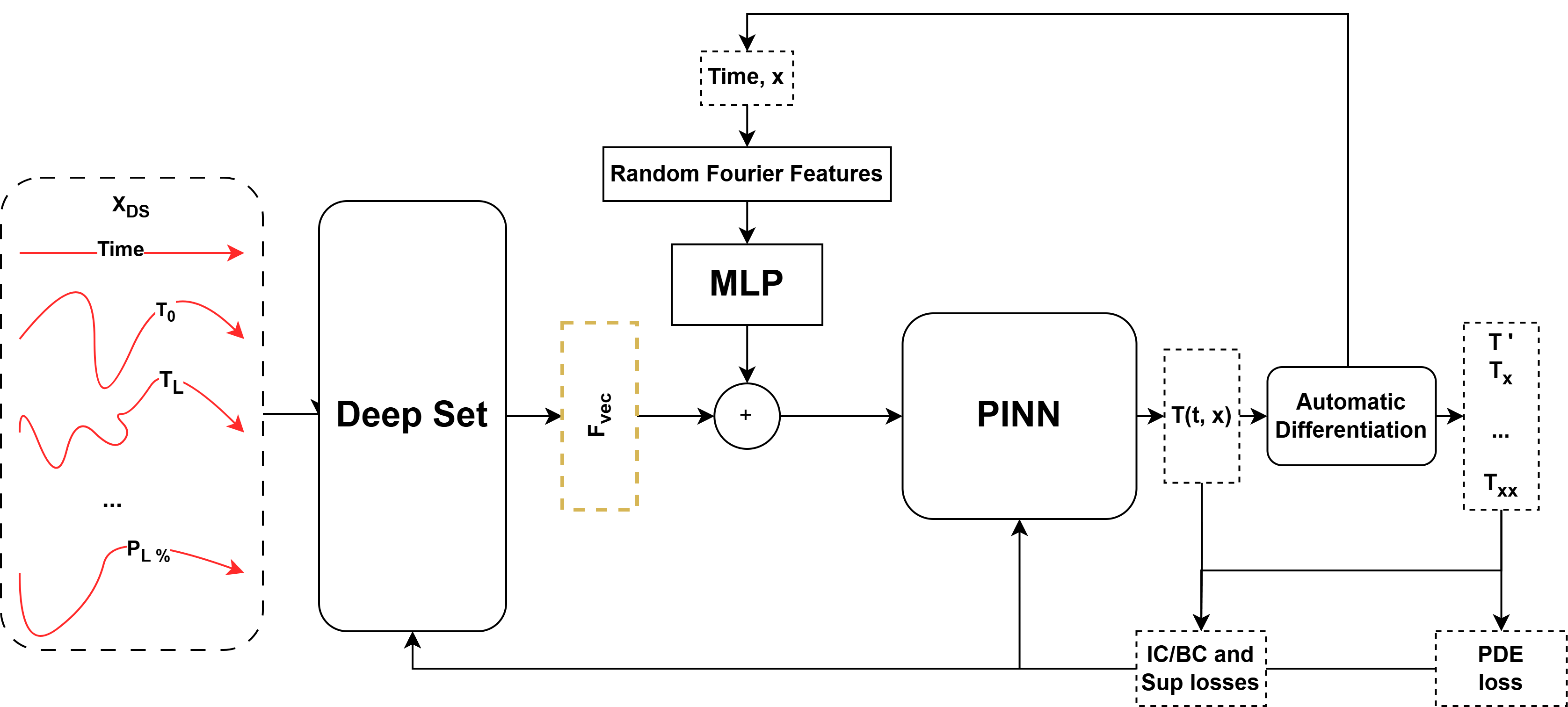}
\caption{The architecture used is designed for the 1D heat application. The Deep Set takes as input $X_{DS}$, which contains sequential data with varying lengths (snapshots) and variable $\Delta T$. It outputs a feature vector, which is then concatenated with a projection of time and space using Random Fourier~\cite{tancik2020fourierfeaturesletnetworks} layer. This concatenated representation is fed into the PINN to predict the temperature. The model is trained using a combination of physics informed loss (PDE), supervised loss (Sup), and initial/boundary condition losses (IC/BC).}
\label{1D_HT_model_v1}
\end{figure}

The architecture used in this study is shown in Fig.~\ref{1D_HT_model_v1}. In this application, the Deep Set takes as input $X_{DS} = [time,~T_0,~T_L,~P_0,~P_L,~L]$, where $\text{time}$ represents the time of each data snapshot. The model is trained with a variable $\Delta T$, which reflects the fact that, in real data, the frequency of acquisition is not constant. We also provide the corresponding temperatures $T_0 = T(t_i, x=0)$ and $T_L = T(t_i, x=L)$ for each time snapshot, as well as the power inputs $P_0 = P(t_i, x=0)$ and $P_L = P(t_i, x=L)$, expressed as percentages (not as heat flux), along with the constant thickness $L$. In this application, we used Random Fourier Features~\cite{tancik2020fourierfeaturesletnetworks} as a projection of the QPs ($time$ and $x$), simply as an encoding step. Since the QPs are identical across all batches during training, the projection is computed once for a single batch and then repeated for the entire batch dimension. This approach helps to reduce redundant computations. This approach helps with the convergence of the model. We used only a PINN, without any identification blocks, as the range of diffusivity is relatively narrow, making the inclusion of additional blocks unnecessary for this application.

\begin{figure}[H]
\centering
\includegraphics[width=0.8\textwidth]{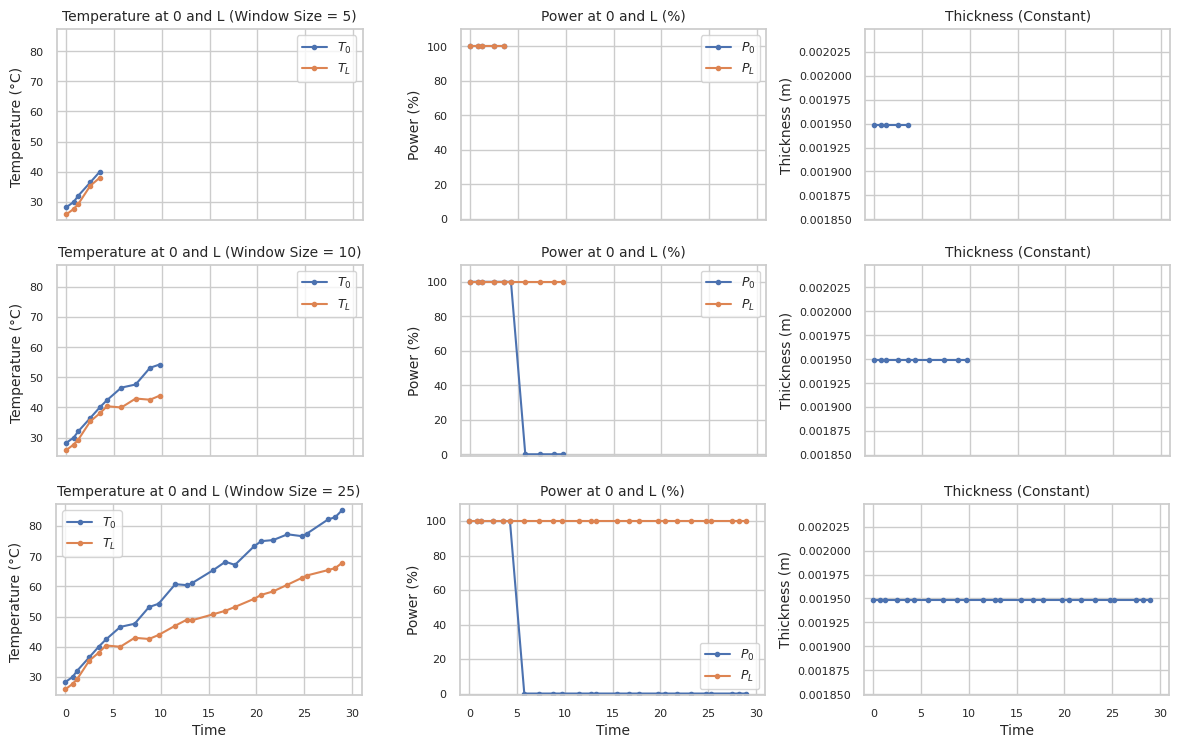}
\caption{The Deep Set input with added white noise, as defined in Eq.~\ref{noiseadded}, with increasing window sizes (sequence lengths) for each row $T \in {5, 10, 25}$, and variable $\Delta T$. On the right, the temperature is shown at the top and bottom ($z \in \{0, L\}$); in the center, the power is given in percentage at $z \in \{0, L\}$; and  the thickness on the right (constant). At time $t = T_{qds} = 5~\text{s}$, the power at the bottom drops to 0\%.}
\label{1D_HT_DS_INP}
\end{figure}

The model is trained solely on synthetic data. We used $N_{\text{train}} = 2200$ samples for training and $N_{\text{val}} = 220$ samples for model selection. The model was then tested on $N_{\text{test}} = 220$ synthetic samples as well as on real data. An example of the model inputs $X_{DS}$ is shown in Fig.~\ref{1D_HT_DS_INP}.
We trained the model using the loss function defined in Eq.~\ref{1D_loss_funct}:

\begin{equation}\label{1D_loss_funct}
\mathcal{L}_{total} = \beta_1 \mathcal{L}_{\text{PDE}} + \beta_2 \mathcal{L}_{\text{IC}} + \beta_3 \mathcal{L}_{\text{BC}} + \beta_4 \mathcal{L}_{\text{SUP}},
\end{equation}
, where the first part, \( \mathcal{L}_{\text{PDE}} \), represents the physics loss (as given in Eq.~\ref{HT-eq}), the second part, \( \mathcal{L}_{\text{IC}} \), represents the initial condition loss, and the third part, \( \mathcal{L}_{\text{BC}} \), represents the boundary condition loss (as given in Eq.~\ref{HT-eq}). The term \( \mathcal{L}_{\text{SUP}} \) represents the supervised loss, which is simply the error between \( T_{\text{real}}(t, x) \) and \( T_{\text{pred}}(t, x) \), for \( x \in \{0, L\} \) and \( t \in [0, T] \). This error varies during the training, both before and after the drop of heat power \( P_{x=L} \) to \( 0 \) (see Fig.~\ref{simu_data_example} and~\ref{1D_HT_DS_INP}). At the beginning of training, the supervised loss is prioritized by initializing the weights as \( \beta_1 = 0 \), \( \beta_2 = 5 \), \( \beta_3 = 1 \), and \( \beta_4 = 10 \). As training progresses, the physics loss weight \( \beta_1 \) is gradually increased once both the supervised and initial condition losses are below specific thresholds. Additionally, the learning rate is adaptively decayed based on the training epochs to ensure stable convergence.

\begin{table}[!h]
\centering
\renewcommand{\arraystretch}{1.3}
\setlength{\tabcolsep}{4pt} 
\caption{Architecture Details of the Proposed Model for Training the 1D Heat Application.}\label{tab:architecture_app3}
\begin{tabular}{|l|p{2cm}|p{1cm}|p{1cm}|p{1cm}|p{5cm}|}
\hline
\textbf{Model} & \textbf{Activation Function} & \textbf{Layers} & \textbf{Width} & \textbf{Output Dim} & \textbf{Comments} \\ 
\hline
Deep Set & $\text{gelu}(X)$ & 3 & 64 & 32 & Aggregation is performed using: \( \sum |.| \). \\ 
\hline
Query Point Encoder & $\text{swish}(X)$ & 3 & 64 & 32 & It takes the projection of time $t$ and space $x$, given by the Random Fourier Layer, as input and maps them to a higher dimensional. \\ 
\hline
PINN & $\text{swish}(X)$ & 3 & 64 & 1 & Takes as input the output of the Deep Set model along with the Query Point Encoder Output. \\ 
\hline
\end{tabular}
\end{table}

\subsubsection{Results}
The model was tested using the test dataset and to ensure that it learned the physics and not just how to predict the temperature at $x=0$ and $x=L$. To verify this, we tested the model's predictions at three other positions. In total we evaluated the model for $x \in \{0,~L/4,~L/2,~3L/4,~L\}$. Since, for real data, we cannot insert sensors inside the plates, we limited the model testing to $x \in \{0,~L\}$. 
For this test, we used the \textbf{MAPE} metric, as there are no outliers or near-zero values, along with the \textbf{MAE} metric. We evaluated the model's predictions over the entire time domain, $t \in [0,~T]$, where $T = 45~\text{s}$, for variable Deep Set input window sizes (Fig.~\ref{1D_HT_DS_INP}). The results for $x \in \{0,~0.5L,~L\}$ are provided in Tab.~\ref{tab:performance_combined}, while results for $x \in \{0.25L,~0.75L\}$ are presented in Tab.~\ref{tab:performance_x25_x75}.
We calculated $\textbf{APE}_{\text{per}=95\%}$ (Absolute Percentage Error at the 95th percentile) and \textbf{MAE} over over the time domain $t \in [0,~T]$ for each test dataset. Then we computed the mean and 90th percentile of these errors over all test datasets, using variable Deep Set input window sizes. The results show that the model predictions are accurate for $x \in [0,~L]$, indicating that the model effectively leveraged the physics to predict temperature within the plate.

\begin{table}[h]
\centering
\renewcommand{\arraystretch}{1.3}
\setlength{\tabcolsep}{6pt}
\caption{Model performance across different spatial locations ($x = 0$, $x = 0.5L$, and $x = L$), using \textbf{MAPE} and \textbf{MAE} metrics. The table summarizes the mean and 90th percentile errors for varying time windows.}
\label{tab:performance_combined}
\begin{tabular}{l|rr|rr|rr|rr|rr|rr}
\toprule
\multirow{2}{*}{\textbf{Time range (s)}} 
& \multicolumn{4}{c|}{$x = 0$} 
& \multicolumn{4}{c|}{$x = 0.5L$} 
& \multicolumn{4}{c}{$x = L$} \\
\cmidrule{2-13}
& \multicolumn{2}{c|}{APE$_{95\%}$ (\%)} & \multicolumn{2}{c|}{MAE} 
& \multicolumn{2}{c|}{APE$_{95\%}$ (\%)} & \multicolumn{2}{c|}{MAE} 
& \multicolumn{2}{c|}{APE$_{95\%}$ (\%)} & \multicolumn{2}{c}{MAE} \\
& Mean & 90\% & Mean & 90\%
& Mean & 90\% & Mean & 90\%
& Mean & 90\% & Mean & 90\% \\
\midrule
0 -- 5  & 4.31 & 8.12 & 2.40 & 4.91 & 4.42 & 8.16 & 2.09 & 4.29 & 4.32 & 8.17 & 2.22 & 4.68 \\
0 -- 10 & 2.69 & 4.57 & 1.27 & 2.53 & 2.90 & 4.54 & 1.11 & 2.30 & 2.98 & 4.85 & 1.11 & 2.24 \\
0 -- 20 & 2.11 & 3.42 & 0.78 & 1.57 & 2.30 & 3.49 & 0.68 & 1.23 & 2.96 & 4.50 & 0.69 & 1.14 \\
0 -- 25 & 2.08 & 3.44 & 0.74 & 1.55 & 2.26 & 3.56 & 0.62 & 1.12 & 2.91 & 4.44 & 0.64 & 1.03 \\
\bottomrule
\end{tabular}
\end{table}

\begin{table}[!h]
\centering
\renewcommand{\arraystretch}{1.3}
\setlength{\tabcolsep}{6pt}
\caption{Model performance for test datasets using \textbf{MAPE} and \textbf{MAE} metrics for \(x = 0.25L\) and \(x = 0.75L\) with variable time windows.}
\label{tab:performance_x25_x75}
\begin{tabular}{l|rrrr|rrrr}
\toprule
\multirow{2}{*}{\textbf{Time (s)}} & \multicolumn{4}{c|}{\textbf{\(x = 0.25L\)}} & \multicolumn{4}{c}{\textbf{\(x = 0.75L\)}} \\
\cmidrule(lr){2-5} \cmidrule(lr){6-9}
 & \multicolumn{2}{c}{APE$_{95\%}$ (\%)} & \multicolumn{2}{c|}{MAE} & \multicolumn{2}{c}{APE$_{95\%}$ (\%)} & \multicolumn{2}{c}{MAE} \\
 & Mean & 90\% & Mean & 90\% & Mean & 90\% & Mean & 90\% \\
\midrule
0 -– 5   & 4.44 & 8.12 & 2.20 & 4.57 & 4.25 & 8.17 & 2.07 & 4.41 \\
0 –- 10  & 2.85 & 4.57 & 1.17 & 2.42 & 2.84 & 4.55 & 1.08 & 2.20 \\
0 –- 20  & 2.19 & 3.48 & 0.71 & 1.38 & 2.37 & 3.51 & 0.66 & 1.19 \\
0 –- 25  & 2.16 & 3.41 & 0.67 & 1.26 & 2.29 & 3.43 & 0.60 & 1.02 \\
\bottomrule
\end{tabular}
\end{table}

We also validated the model on real data. The results for $x\in {0,~L}$ are given using the same metrics in Tab.~\ref{tab:combined_performance_exp} for the two PA66 plates and for the ten PP plates. The results are presented with variable window sizes (number of points given as input to the Deep Set), since for real data, calculating the metrics for all data (grouped by time) is not possible due to the variable acquisition frequency. Overall, the results are very satisfying. 

\begin{table}[h]
\centering
\renewcommand{\arraystretch}{1.3}
\setlength{\tabcolsep}{5.5pt}
\caption{Model performance on real data using \textbf{MAPE} and \textbf{MAE} metrics for PA66 and PP plates at positions \(x = 0\) and \(x = L\), across different window sizes.}
\label{tab:combined_performance_exp}
\begin{tabular}{l|rrrr|rrrr}
\toprule
\multirow{2}{*}{\textbf{Window}} & \multicolumn{4}{c|}{\textbf{PA66-plate} \(x = 0\)} & \multicolumn{4}{c}{\textbf{PP-plate} \(x = 0\)} \\
& \multicolumn{2}{c}{APE$_{95\%}$ (\%) } & \multicolumn{2}{c|}{MAE } & \multicolumn{2}{c}{APE$_{95\%}$ (\%) } & \multicolumn{2}{c}{MAE } \\
\textbf{Size} & Mean & 90\% & Mean & 90\% & Mean & 90\% & Mean & 90\% \\
\midrule
5  & 24.50 & 26.87 & 7.46 & 8.14 & 13.33 & 18.21 & 4.34 & 5.98 \\
10 & 5.96  & 6.14  & 1.50 & 1.96 & 6.98  & 11.84 & 2.20 & 3.34 \\
15 & 5.07  & 6.15  & 1.13 & 1.14 & 4.51  & 6.23  & 1.63 & 2.49 \\
20 & 5.21  & 6.29  & 1.15 & 1.16 & 4.12  & 6.45  & 1.31 & 1.96 \\
25 & 5.30  & 6.37  & 1.20 & 1.20 & 4.10  & 6.14  & 1.18 & 1.74 \\
30 & 5.44  & 6.42  & 1.25 & 1.26 & 4.10  & 5.81  & 1.05 & 1.52 \\
\midrule
& \multicolumn{4}{c|}{\textbf{PA66-plate \(x = L\)}} & \multicolumn{4}{c}{\textbf{PP-plate \(x = L\)}} \\
\textbf{Size} & Mean & 90\% & Mean & 90\% & Mean & 90\% & Mean & 90\% \\
\midrule
5  & 40.52 & 42.61 & 12.25 & 12.79 & 49.28 & 62.38 & 15.34 & 20.45 \\
10 & 4.33  & 4.64  & 1.07  & 1.29  & 21.99 & 47.41 & 6.06  & 13.57 \\
15 & 3.17  & 3.28  & 0.50  & 0.55  & 4.62  & 6.94  & 1.10  & 1.63  \\
20 & 3.01  & 3.40  & 0.43  & 0.47  & 4.32  & 6.31  & 1.03  & 1.58  \\
25 & 2.98  & 3.38  & 0.43  & 0.47  & 4.39  & 6.25  & 1.03  & 1.57  \\
30 & 2.94  & 3.24  & 0.42  & 0.47  & 4.68  & 6.33  & 1.06  & 1.59  \\
\bottomrule
\end{tabular}
\end{table}

\section{Analysis and conclusion}
Based on these results, the combination of Deep set and PINN is a very promising approach. Deep set is highly effective in encoding dynamic behavior and mapping it to dynamic parameters, as shown by~\cite{SEQENCODER}. By combining it with PINN, we can encode variable changes in boundary conditions, initial conditions, or dynamic parameters, and map the encoded vector to the dynamic behavior. We demonstrate that we can include a physics loss directly in the training process. Furthermore, we have shown that adding additional blocks can be used to cross-validate the model's predictions and provide a confidence level for the model's output. Using Deep set allows us to handle sequences of variables with varying time steps and varying lengths, noisy data, and can also encode variable geometries. \\
Nevertheless, for example, in the case of the 2D Navier-Stokes equations, incorporating multiple geometry changes as inputs to the Deep Set can be challenging. For these cases, we believe that additional encoding blocks would be necessary. In cases where available data are 2D spatial fields or images, it is important to add special image encoding networks, such as CNNs. In fact, the Set Transformer, which has a similar principle as Deep set, has been tested in combination with CNNs and shown very promising results. Combining these networks with additional blocks will allow for the use of different data formats as input. For future work, we will test the ability to combine this approach with an image sequence encoder and combine it with PINN to extrapolate in time using the physics.

\bibliography{main}

\end{document}